\documentclass{clv3web}

\usepackage{hyperref}
\usepackage[table]{xcolor}
\usepackage{multirow}
\usepackage{makecell}
\definecolor{darkblue}{rgb}{0, 0, 0.5}
\hypersetup{colorlinks=true,citecolor=darkblue, linkcolor=darkblue, urlcolor=darkblue}

\newcommand{\eqto}{\phantom{{}+{}}=\phantom{{}+{}}}

\issue{1}{1}{2016}

\runningtitle{Crosslingual Transfer in Topic Modeling}
\runningauthor{Hao and Paul}
\author{Shudong Hao}
\affil{University of Colorado Boulder}

\author{Michael J. Paul}
\affil{University of Colorado Boulder}

\newcommand{\vt}[1]{\mathbf{#1}}

\newcommand{\ie}{\textit{i.e., }}

\newcommand{\sv}[0]{\textsc{sv}}
\newcommand{\de}[0]{\textsc{de}}
\newcommand{\es}[0]{\textsc{es}}
\newcommand{\ar}[0]{\textsc{ar}}

\newcommand{\zh}[0]{\textsc{zh}}
\newcommand{\ru}[0]{\textsc{ru}}
\newcommand{\en}[0]{\textsc{en}}

\newcommand{\am}[0]{\textsc{am}}
\newcommand{\ay}[0]{\textsc{ay}}

\newcommand{\mk}[0]{\textsc{mk}}

\newcommand{\sw}[0]{\textsc{sw}}
\newcommand{\tl}[0]{\textsc{tl}}
\mathchardef\mhyphen="2D

\newcommand{\highlan}[0]{\textsc{HighLan}}
\newcommand{\lowlan}[0]{\textsc{LowLan}}
\newcommand{\softlink}[0]{\textsc{softlink}}
\newcommand{\voclink}[0]{\textsc{voclink}}
\newcommand{\doclink}[0]{\textsc{doclink}}

\newcommand{\probilda}[0]{\textsc{ProbBiLDA}}
\newcommand{\cbilda}[0]{\textsc{C-BiLDA}}
\newcommand{\mlhptlm}[0]{\textsc{mlhPLTM}}

\newcommand{\ted}[0]{\textsc{ted}}
\newcommand{\gv}[0]{\textsc{gv}}
\newcommand{\cnpmi}[0]{\textsc{cnpmi}}
\newcommand{\npmi}[0]{\textsc{npmi}}

\newcommand{\hl}{\cellcolor{gray!10}}

\newcommand{\eq}[1]{Equation~(\ref{#1})}
\newcommand{\fig}[1]{Figure~\ref{#1}}
\renewcommand{\tbl}[1]{Table~\ref{#1}}

\usepackage{ragged2e}
\usepackage{graphicx}
\usepackage{amsmath,amssymb,amsthm,dsfont,color,enumitem}
\usepackage{url}
\usepackage{subcaption}
\usepackage{bold-extra}
\usepackage{bm}
\usepackage{xcolor}
\usepackage{comment}
\usepackage[ruled,vlined,noend,linesnumbered]{algorithm2e}

\allowdisplaybreaks

\title{An Empirical Study on Crosslingual Transfer in Probabilistic Topic Models}

\begin{document}
	\maketitle
	\begin{abstract}
		Probabilistic topic modeling is a common first step in crosslingual tasks to enable
		knowledge transfer and extract multilingual features.
		While many multilingual topic models have been developed,
		their assumptions about the training corpus are quite varied,
		and it is not clear how well the different models can be utilized
		under various training conditions.
		In this paper,
		we systematically study the knowledge transfer mechanisms behind different multilingual topic models,
		and through a broad set of experiments with four models on ten languages, we provide empirical insights that can inform
		the selection and future development of multilingual topic models.
	\end{abstract}

	\section{Introduction}
	Popularized by Latent Dirichlet Allocation~\cite{BleiNJ03},
	probabilistic topic models have been an important tool for analyzing large collections of texts~\cite{Blei12,Blei18}.
	Their simplicity and interpretability
	make topic models popular for many natural language processing tasks,
	such as discovering document networks~\cite{ChenZXZ13,ChangB09} and
	authorship attribution~\cite{SeroussiZB14}.

	Topic models take a corpus $D$ as input,
	where each document $d\in D$ is usually represented as a sparse vector in a vocabulary space,
	and project these documents to a lower-dimensional \textit{topic space}.
	In this sense,
	topic models are often used as a dimensionality reduction technique
	to extract representative and human-interpretable features.
	
\begin{figure*}
		\centering
		\begin{subfigure}[t]{0.5\linewidth}
			\centering
			\includegraphics[width=0.6\textwidth]{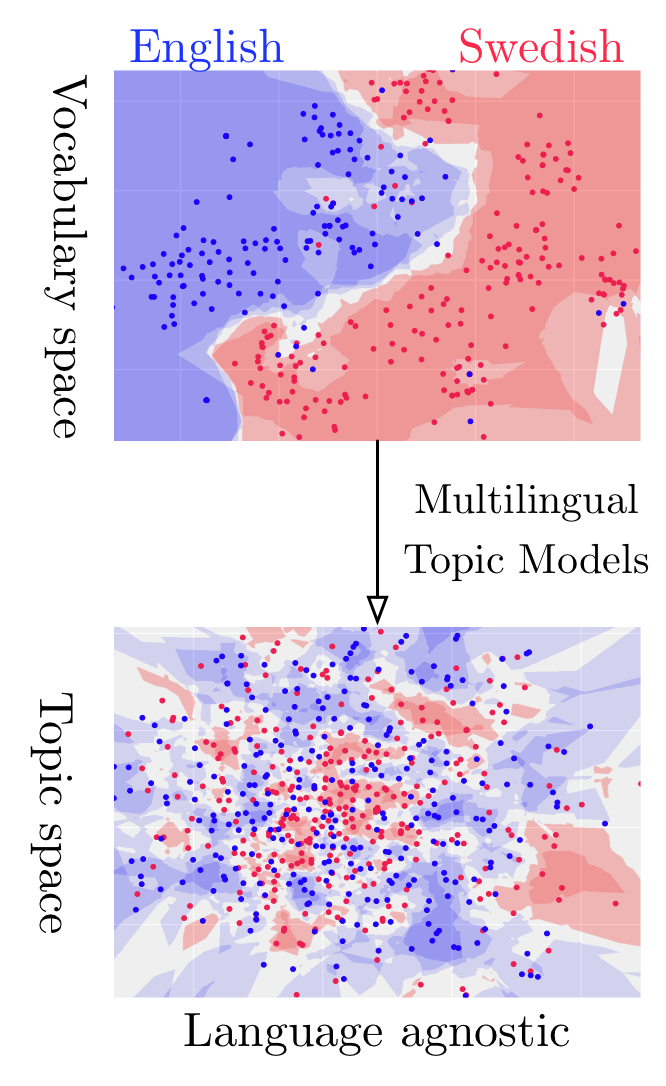}
			\caption{Multilingual topic models project language-specific and high-dimensional features from the vocabulary space to a language-agnostic and low-dimensional topic space. This figure shows a t-SNE~\cite{maaten2008visualizing} representation of a real dataset.}
			\label{fig:topicspace}
		\end{subfigure}~
		\begin{subfigure}[t]{0.5\linewidth}
			\centering
			\includegraphics[width=\linewidth]{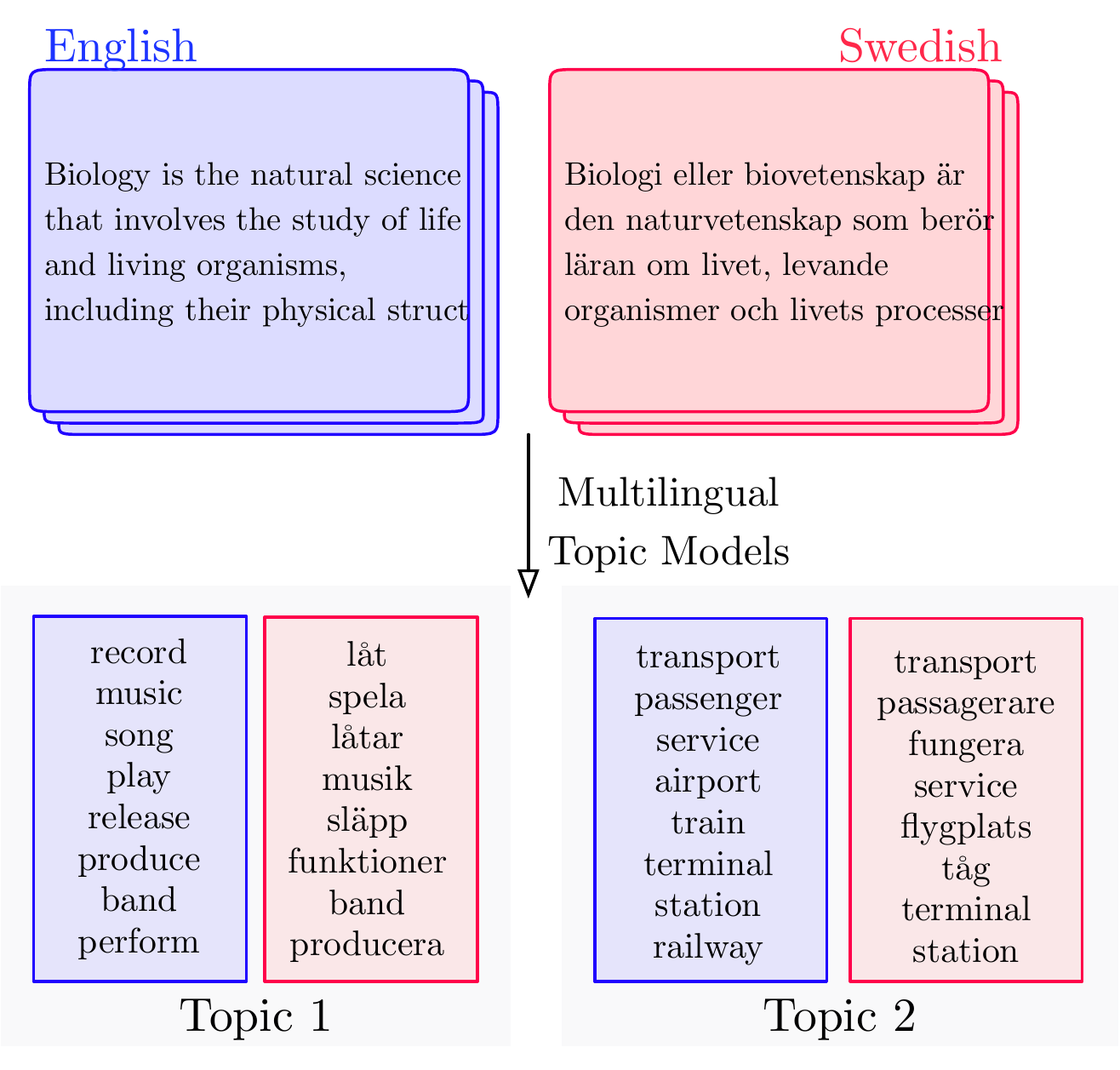}
			\caption{Multilingual topic models produce theme-aligned topics for all languages. From a human's view, each topic contains  different languages but the words are describing the same thing.}
			\label{fig:corpustotopic}
		\end{subfigure}
		\caption{Overview of multilingual topic models.}
	\end{figure*}

	Text collections, however, are often not in a single language,
	and thus there has been a need to generalize topic models from monolingual to multilingual settings.
	Given a corpus $D^{(1,\ldots,L)}$ in languages $\ell \in \{1,\ldots, L\}$,
	multilingual topic models learn topics in each of the languages.
	From a human's view,
	each topic should be related to the same theme,
	even if the words are not in the same language  (\fig{fig:corpustotopic}).
	From a machine's view,
	the word probabilities within a topic should be similar across languages, such that the low-dimensional representation of documents is not dependent on the language.
	In other words, the topic space in multilingual topic models is \textit{language agnostic} (\fig{fig:topicspace}).
	
	This paper presents two major contributions to multilingual topic models.
	We first provide an alternative view of multilingual topic models
	by explicitly formulating a crosslingual knowledge transfer process 
	during posterior inference (Section~\ref{sec:transfer}).
	Based on this analysis,
	we unify different multilingual topic models by defining a function called the {\em transfer operation}. This function provides an abstracted view of the knowledge transfer mechanism behind these models, while enabling further generalizations and improvements. Using this formulation, we analyze several existing multilingual topic models (Section~\ref{sec:models}).

	Second, in our experiments
	we compare four representative models under
	different training conditions (Section~\ref{sec:expr}).
	The models are trained and evaluated on ten languages from various language families to increase language diversity in the experiments.
	In particular, we include five languages with relatively high resources and the others with low resources.
	To quantitatively evaluate the models,
	we focus on topic quality in Section~\ref{sec:in},
	and performance of downstream tasks using crosslingual document classification in Section~\ref{sec:ex}.
	We investigate how sensitive the models are to different language resources (\textit{i.e.,} parallel/comparable corpus and dictionaries), and analyze what factors cause this difference (Sections~\ref{ex:ex} and \ref{wt}).

	\section{Background}
	
	We first review monolingual topic models, focusing on Latent Dirichlet Allocation,
	and then describe two families of multilingual extensions.
	Based on the types of supervision added to multilingual topic models,
	we separate the two model families into document-level and word-level supervision.
	
	Topic models provide a high-level view of latent thematic structures in a corpus.
	Two main branches for topic models are non-probabilistic approaches such as Latent Semantic Analysis (LSA, \namecite{DeerwesterDLFH90}) and Non-Negative Matrix Factorization (NMF, \namecite{XuLG03}),
	and probabilistic ones such as Latent Dirichlet Allocation (LDA, \namecite{BleiNJ03}) and probabilistic LSA (pLSA, \namecite{Hofmann99}).
	All these models were originally developed for monolingual data and later adapted to multilingual situations.
	Though there has been work to adapt non-probabilistic models, for example based on ``pseudo-bilingual'' corpora approaches~\cite{littman1998automatic},
	most multilingual topic models that are trained on multilingual corpora 
	are based on probabilistic models, especially LDA.
	Therefore,
	our work is focused on the probabilistic topic models,
	and in the following section
	we start by describing LDA.
	
	\subsection{Monolingual Topic Models}
	
\begin{figure}
		\centering
		\includegraphics[width=0.7\linewidth]{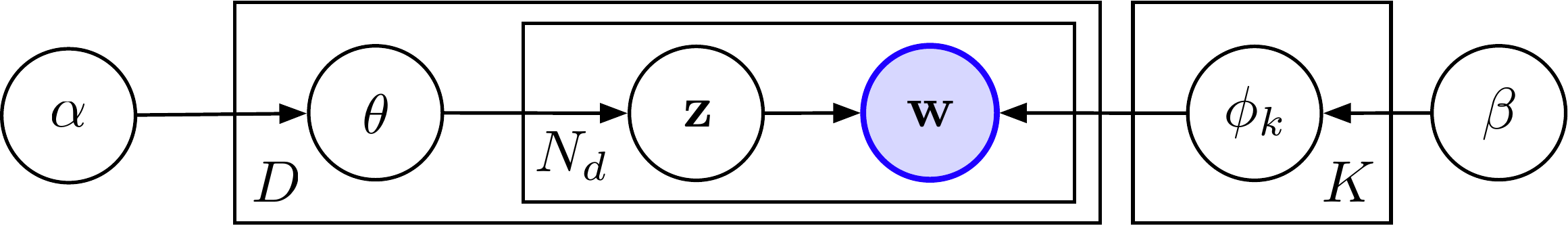}
		\caption{Plate notation of LDA. $\alpha$ and $\beta$ are Dirichlet hyperparameters for $\theta$ and $\{\phi^{(k)}\}_{k=1}^K$. Topic assignments are denoted as $\mathbf{z}$, and $\mathbf{w}$ denotes observed tokens.}
		\label{fig:ldaplate}
	\end{figure}

	The most popular topic model is Latent Dirichlet Allocation (LDA),
	introduced by~\namecite{BleiNJ03}.
	This model assumes each document $d$ is represented by
	a multinomial distribution $\theta_{d}$ over topics,
	while each ``topic'' $k$ is a multinomial distribution
	$\phi^{(k)}$ over the vocabulary $V$.
	In the generative process,
	each $\theta$ and $\phi$ are generated from Dirichlet distributions
	parameterized by $\alpha$ and $\beta$, respectively.
	The hyperparameters for Dirichlet distributions
	can be asymmetric~\cite{WallachMM09}, though in this work we use symmetric priors.
	Figure~\ref{fig:ldaplate} shows the plate notation of \textsc{lda}.

	\subsection{Multilingual Topic Models}
	
	We now describe a variety of multilingual topic models, organized into two families based on the type of supervision they use.
	Later in Section~\ref{sec:models},
	we focus on a subset of the models described here for deeper analysis using our knowledge transfer formulation,
	selecting the most general and representative models.

	\subsubsection{Document Level}
	The first model proposed to process multilingual corpora using LDA
	is the Polylingual Topic Model (PLTM, \namecite{MimnoWNSM09}; \namecite{NiSHC09}).
	This model extracts language-consistent topics from parallel
	or highly comparable multilingual corpora (for example, Wikipedia articles aligned across languages),
	assuming that document translations share the same topic distributions.
	This model has been extensively used and adapted
	in various ways for different crosslingual tasks~\cite{KrstovskiS16,LiuDM15,VulicM14,MoensV13,KrstovskiS11}.
	
	In the generative process,
	PLTM first generates language-specific topic-word distributions $\phi^{(\ell,k)}\sim\mathrm{Dir}\left(\beta^{(\ell)}\right)$,
	for topics $k=1,\ldots,K$ and languages $\ell=1,\ldots,L$.
	Then, for each document tuple $\mathbf{d}=\left(d^{(1)},\ldots,d^{(L)}\right)$,
	it generates a \textit{tuple}-topic distribution $\theta_{\mathbf{d}}\sim\mathrm{Dir}(\alpha)$.
	Every topic in this document tuple is generated from $\theta_{\mathbf{d}}$,
	and the word tokens in this document tuple are then generated from language-specific word distributions $\phi^{(\ell,k)}$ for each language.
	To apply PLTM,
	the corpus  must be parallel or closely comparable
	to provide document-level supervision.
	We refer to this as the \textbf{document links} model (\textbf{\doclink{}}).
	
	Models that transfer knowledge on the document level
	have many variants,
	including
	\textbf{\softlink{}}~\cite{HaoP18},
	comparable bilingual latent Dirichlet allocation (\textbf{\cbilda{}}, \namecite{HeymanVM16}),
	the partially-connected multilingual topic model (\textbf{\textsc{pcMLTM}}, \namecite{LiuDM15}),
	and multi-level hyperprior polylingual topic model (\textbf{\mlhptlm{}}, \namecite{KrstovskiSK16}).
	\softlink{} generalizes \doclink{} by using a dictionary,
	so that documents can be linked based on overlap in their vocabulary, even if the corpus is not parallel or comparable.
	\cbilda{} is a direct extension of \doclink{}
	which also models language-specific distributions
	to distinguish topics that are shared across languages from language-specific topics.
	\textsc{pcMLTM} adds an additional observed variable
	to indicate the absence of a language in a document tuple.
	\mlhptlm{} uses a hierarchy of hyperparameters to generate section-topic distributions.
	This model was motivated by applications to scientific research articles,
	where each section $s$ has its own topic distribution $\theta^{(s)}$ shared by both languages.
	
	\subsubsection{Word Level}
	
	Instead of document-level connections between languages,
	\namecite{BoydGraberB09} and~\namecite{JagarlamudiD10} proposed to
	model connections between languages through {\em words} using a multilingual dictionary and apply hyper-Dirichlet Type-I distributions~\cite{AndrzejewskiZC09,dennis1991}.
	We refer to these approaches as the \textbf{vocabulary links} model (\textbf{\voclink{}}).

\begin{figure}\centering
		\includegraphics[width=0.8\linewidth]{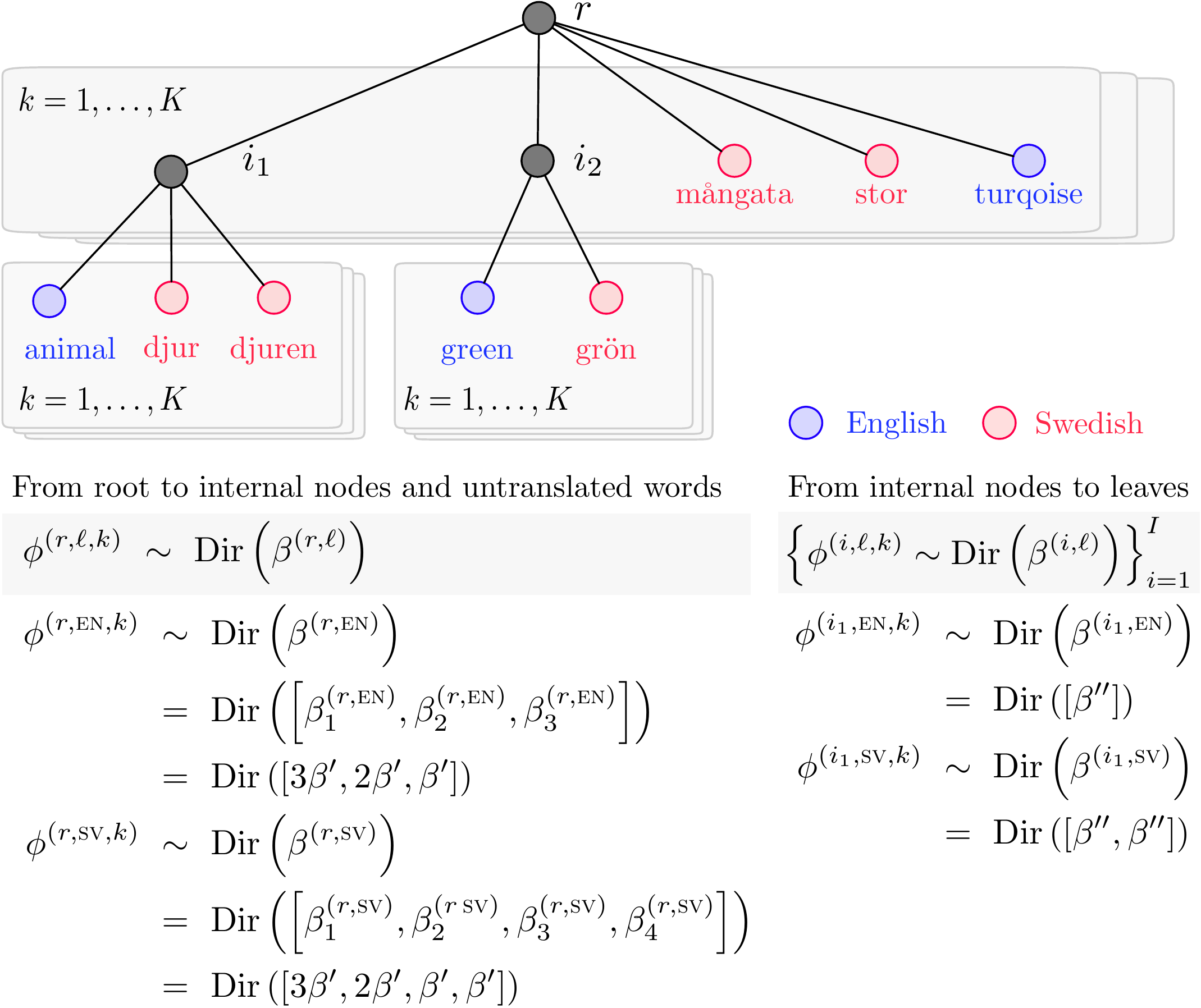}
		\caption{An illustration of the tree structure used in word-level models. Hyperparameters $\beta^{(r,\ell)}$ and $\beta^{(i,\ell)}$ are both vectors, while $\beta'$ and $\beta''$ are scalars.
			In the figure, $i_1$ has three translations, so the corresponding hyperparameter $\beta^{(r,\en{})}_1 = \beta^{(r,\sv{})}_1 = 3\beta'$.}
		\label{tree}
	\end{figure}

	Specifically,
	\voclink{} uses a dictionary to create a tree structure
	where each internal node contains word translations,
	and words that are not translated are attached directly to the root of the tree $r$ as leaves.
	In the generative process,
	for each language $\ell$,
	\voclink{} first generates $K$ multinomial distributions over all internal nodes and word types that are not translated,
	$\phi^{(r,\ell,k)}\sim\mathrm{Dir}\left(\beta^{(r,\ell)}\right)$,
	where $\beta^{(r,\ell)}$ is a vector of Dirichlet prior from root $r$ to
	internal nodes and untranslated words in language $\ell$.
	Then under each internal node $i$,
	for each language $\ell$,
	\voclink{} generates a multinomial $\phi^{(i,\ell,k)}\sim\mathrm{Dir}\left(\beta^{(i,\ell)}\right)$
	over word types in language $\ell$ under the node $i$.
	Note that both $\beta^{(r,\ell)}$ and $\beta^{(i,\ell)}$ are vectors.
	In the first vector $\beta^{(r,\ell)}$, each cell is parameterized by scalar $\beta'$
	and scaled by the number of word translations under that internal node.
	For the second vector $\beta^{(i,\ell)}$,
	it's a symmetric hyperparameter where every cell uses the same scalar $\beta''$.
	See Figure~\ref{tree} for an illustration.
	
	Thus,
	to draw a word in language $\ell$ is equivalent to generating a \textit{path} from the root to leaf nodes, \textit{i.e.,} $\left(r\rightarrow i, i \rightarrow w^{(\ell)}\right)$ or 
	$\left(r \rightarrow w^{(\ell)}\right)$:
	\begin{align}
		\Pr\left(r\rightarrow i, i \rightarrow w^{(\ell)} | k\right) &\eqto \Pr\left(i | k\right) \cdot \Pr\left(w^{(\ell)} | k, i\right),\\
		\Pr\left(r\rightarrow w^{(\ell)} | k\right) &\eqto \Pr\left(w^{(\ell)} | k\right).
	\end{align}
	Document-topic distributions $\theta_{d}$
	are generated in the same way
	as monolingual LDA,
	since no document translation is required.
	
	The use of dictionaries to model similarities across topic-word distributions
	has been formulated in other ways as well.
	\textbf{\probilda{}} \cite{MaN17} 
	uses inverted indexing~\cite{SogaardAAPBJ15}
	to encode assumptions that word translations
	are generated from same distributions.
	\probilda{} does not use
	tree structures in the parameters as in \voclink{},
	but the general idea of sharing distributions among word translations
	is similar.	
	\namecite{GutierrezSLMG16} use part-of-speech taggers
	to separate topic words (nouns) and perspective words (adjectives and verbs),
	developed for the application of detecting cultural differences, \ie 
	how different languages have different perspectives on the same topic.
	Topic words are modeled in the same way as in \voclink{},
	while perspective words are modeled in a monolingual fashion.
	
	\section{Crosslingual Transfer in Probabilistic Topic Models}
	\label{sec:transfer}

	Conceptually, the term ``knowledge transfer'' indicates that
	there's a \textit{process} of carrying information from a source to a destination.
	Using the representations of graphical models,
	the process can be visualized as the dependence of random variables.
	For example,
	$X\rightarrow Y$ implies that the generation of variable $Y$ is conditioned on $X$,
	and thus the information of $X$ is carried to $Y$.
	If $X$ represents a probability distribution,
	the distribution of $Y$ is informed by $X$,
	presenting a process of knowledge transfer,
	as we define it in this work.
	
	In our study,
	``knowledge'' can be loosely defined as $K$ multinomial distributions over the vocabularies,
	\textit{i.e.,} $\{\phi^{(k)}\}_{k=1}^K$.
	Thus, to study the transfer mechanisms in topic models
	is to reveal how the models transfer $\{\phi^{(k)}\}_{k=1}^K$ from one language to another.
	To date,
	this transfer process has not been obvious in most models,
	because typical multilingual topic models assume the tokens in multiple languages are generated jointly.

	In this section, 
	we present a reformulation of these models that 
	breaks down the co-generation assumption of current models
	and instead explicitly show the dependencies between languages.
	Starting with a simple example in Section~\ref{sec:expose},
	we show that our alternative formulation 
	derives the same collapsed Gibbs sampler, and thus the same posterior distribution over samples,
	as in the original model.
	With this prerequisite, in Section~\ref{sec:to}
	we introduce the {\em transfer operation},
	which will be used to generalize and extend current multilingual topic models in Section~\ref{sec:models}.

	\subsection{Transfer Dependencies}\label{sec:expose}

	\begin{figure*}[t!]
		\centering
		\begin{subfigure}[t]{0.32\textwidth}
			\includegraphics[width=\linewidth]{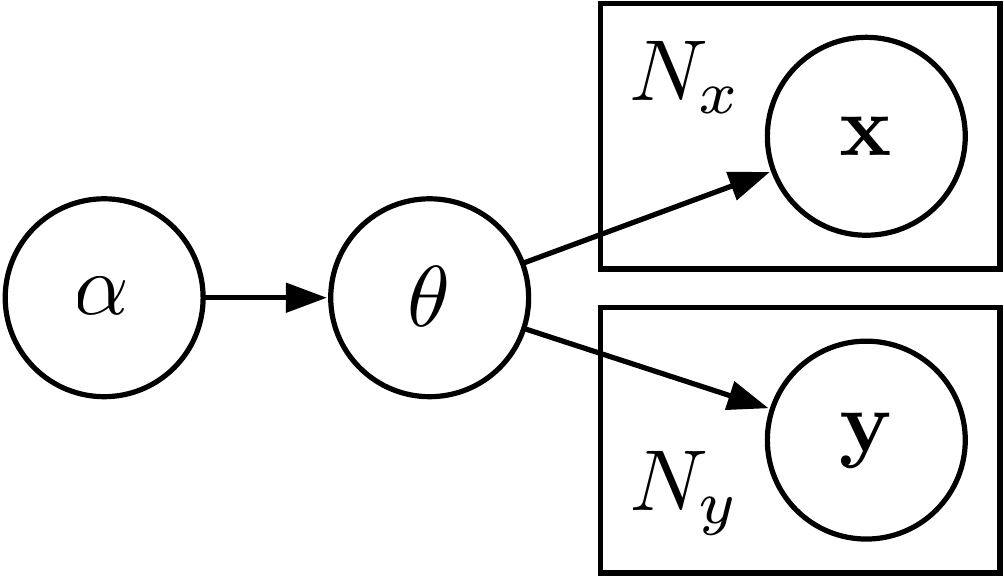}
			\caption{}
			\label{cogen1}
		\end{subfigure}~
		\begin{subfigure}[t]{0.32\textwidth}
			\includegraphics[width=\linewidth]{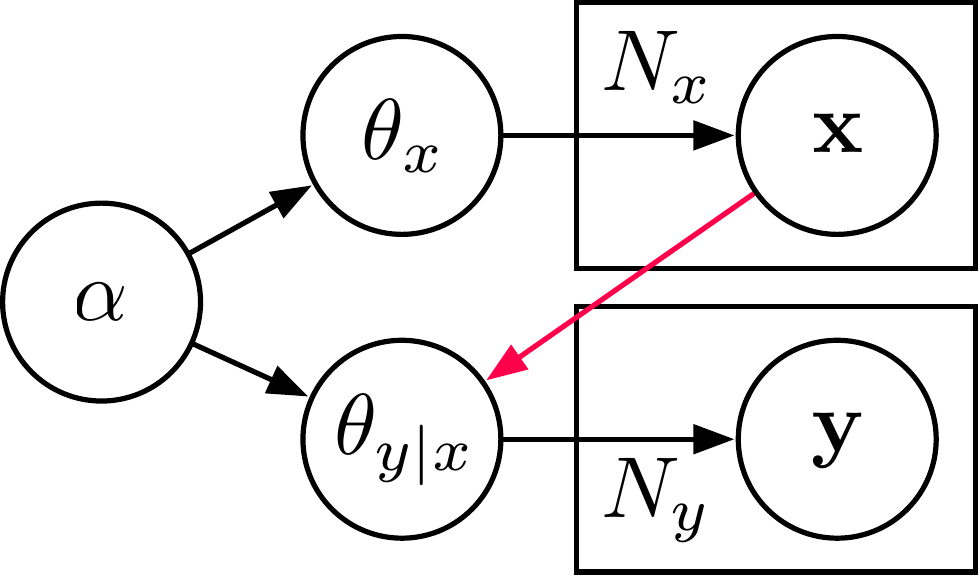}
			\caption{}
			\label{cogen2}
		\end{subfigure}~
		\begin{subfigure}[t]{0.32\textwidth}
			\includegraphics[width=\linewidth]{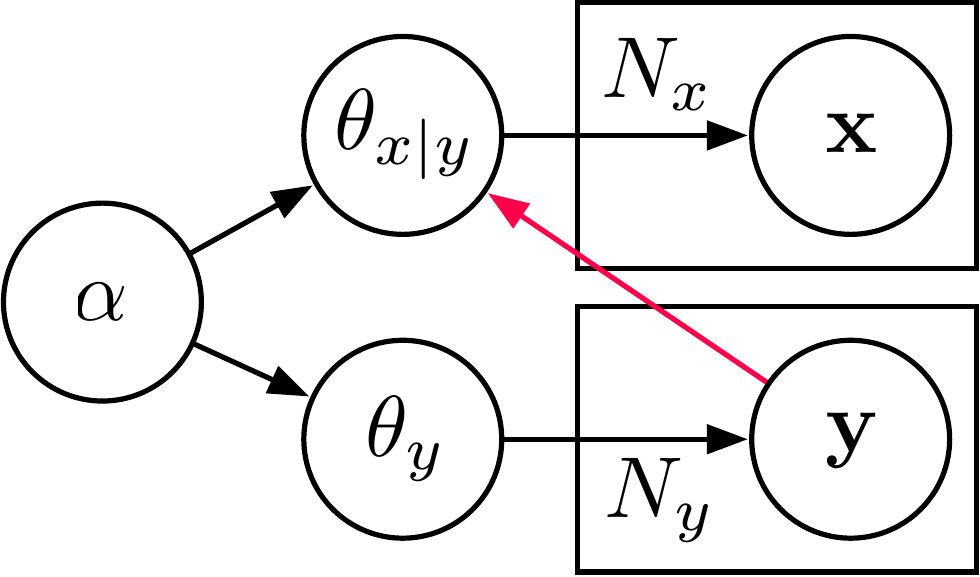}
			\caption{}
			\label{cogen3}
		\end{subfigure}
		\caption{The co-generation assumption generates $\mathbf{x}$ and $\mathbf{y}$ at the same time from the same $\theta$~(Figure~\ref{cogen1}). To make the transfer process clear, we make the generation of $\mathbf{y}$ conditional on $\mathbf{x}$~(Figure~\ref{cogen2}) and highlight the dependency in red. Since both $x$ and $y$ are exchangeable,
			the dependency can go the other way as shown in Figure~\ref{cogen3}.}
	\end{figure*}

	We start with a simple graphical model, where
	$\theta\in\mathbb{R}_+^K$ is a $K$-dimensional categorical distribution,
	drawn from a Dirichlet parameterized by $\alpha$, a symmetric hyperparameter (Figure~\ref{cogen1}).
	Using $\theta$,
	the model generates two variables, $X$ and $Y$,
	and we use $\mathbf{x}$ and $\mathbf{y}$ to denote the generated observations.
	In the co-generation assumption,
	the variables $X$ and $Y$ are generated from the same $\theta$ at the same time,
	without dependencies between each other.
	Thus, we call this the {joint model} denoted as $\mathcal{G}^{(X,Y)}$ 
	and the probability of the sample $(\mathbf{x},\mathbf{y})$ is
	${\Pr}\left( \mathbf{x}, \mathbf{y} ; \alpha, \mathcal{G}^{(X,Y)}\right)$.

	According to Bayes' theorem,
	there are two equivalent ways to expand the probability of $(\mathbf{x},\mathbf{y})$:
	\begin{align}
		{\Pr}\left( \mathbf{x}, \mathbf{y} ; \alpha\right) ~=~& \Pr\left( \mathbf{x}| \mathbf{y}; \alpha\right) \cdot \Pr\left(\mathbf{y};\alpha\right),\\
		{\Pr}\left( \mathbf{x}, \mathbf{y} ; \alpha\right) ~=~& \Pr\left( \mathbf{y}| \mathbf{x}; \alpha\right) \cdot \Pr\left(\mathbf{x};\alpha\right),
	\end{align}
	where we notice that the generated sample is conditioned on another sample,
	\textit{i.e.,} $\Pr\left( \mathbf{x}| \mathbf{y}; \alpha\right)$ and $\Pr\left( \mathbf{y}| \mathbf{x}; \alpha\right)$,
	which fits into our concept of ``transfer''.
	We show both cases in Figures~\ref{cogen2} and \ref{cogen3},
	and denote the graphical structures as $\mathcal{G}^{(Y|X)}$ and $\mathcal{G}^{(X|Y)}$, respectively,
	to show the dependencies between the two variables.
	
	In this formulation,
	the model generates $\theta_x$ from $\mathrm{Dirichlet}(\alpha)$ first
	and uses $\theta_x$ to generate the sample of $\mathbf{x}$.
	Using the histogram of $\mathbf{x}$ denoted as $\mathbf{n}_x=[n_{1|x},n_{2|x},\ldots,n_{K|x}]$
	where $n_{k|x}$ is the number of instances of $X$ assigned to category $k$,
	together with hyperparameter $\alpha$,
	the model then generates a categorical distribution $\theta_{y|x}\sim\mathrm{Dir}(\mathbf{n}_x+\alpha)$,
	from which the sample $\mathbf{y}$ is drawn.
	
	This differs from the original joint model in that original parameter vector $\theta$ has been replaced with two variable-specific parameter vectors.
	The next section derives posterior inference with Gibbs sampling after integrating out the $\theta$ parameters, and we show that the sampler for each of two model formulations is equivalent and thus samples from an equivalent posterior distribution over $\mathbf{x}$ and $\mathbf{y}$.

	\subsection{Collapsed Gibbs Sampling}\label{sec:gs}
	
	General approaches to infer posterior distributions over graphical model variables
	include Gibbs sampling, variational inference, and hybrid approaches~\cite{KimVS13}.
	We focus on collapsed Gibbs sampling~\cite{griffiths2004finding},
	which marginalizes out the parameters ($\theta$ in the example above) to focus on the variables of interest ($\mathbf{x}$ and $\mathbf{y}$ in the example).

	\begin{figure*}
		\begin{subfigure}[t]{0.5\textwidth}
			\includegraphics[width=\linewidth]{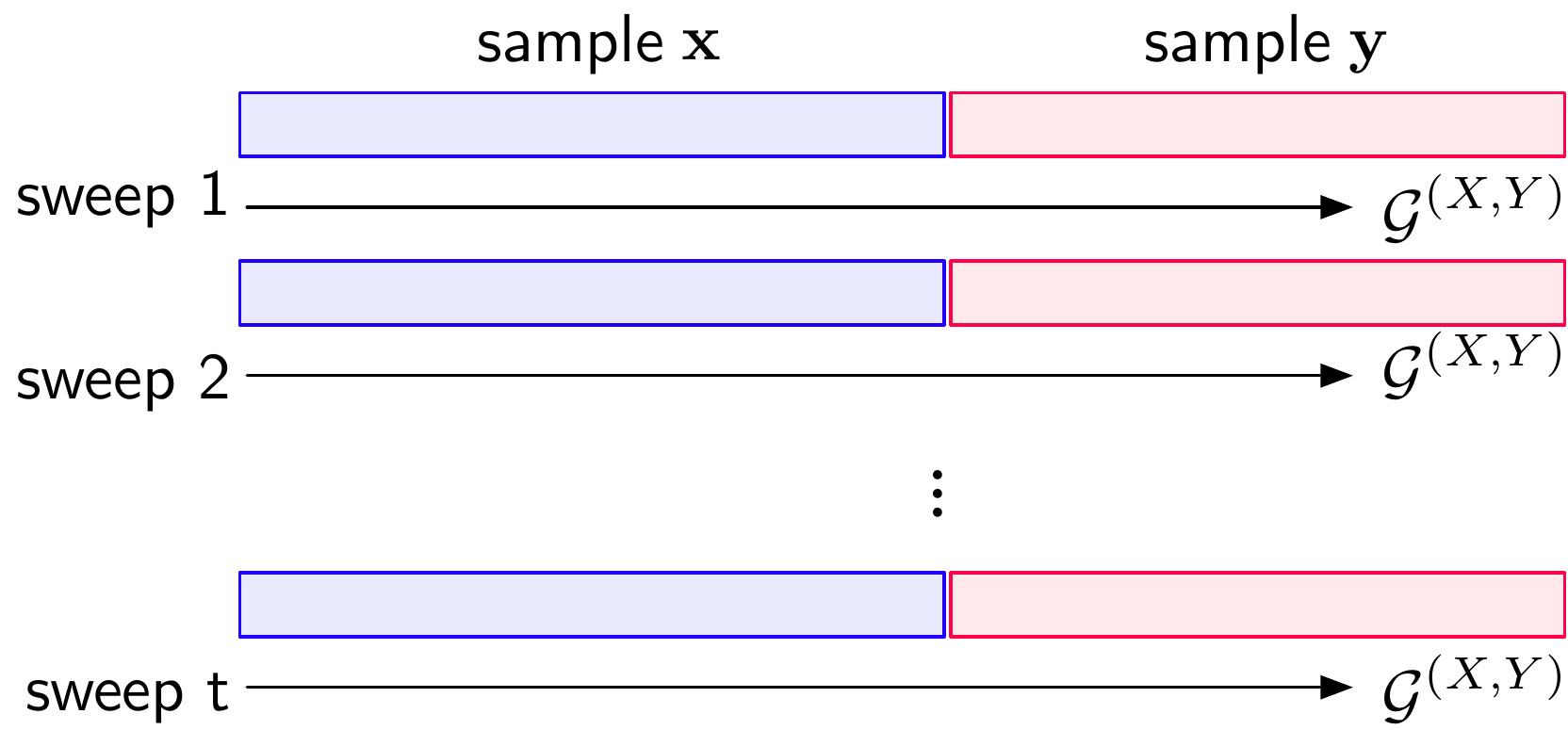}
			\caption{}
			\label{gibbsco}
		\end{subfigure}~
		\begin{subfigure}[t]{0.5\textwidth}
			\includegraphics[width=\linewidth]{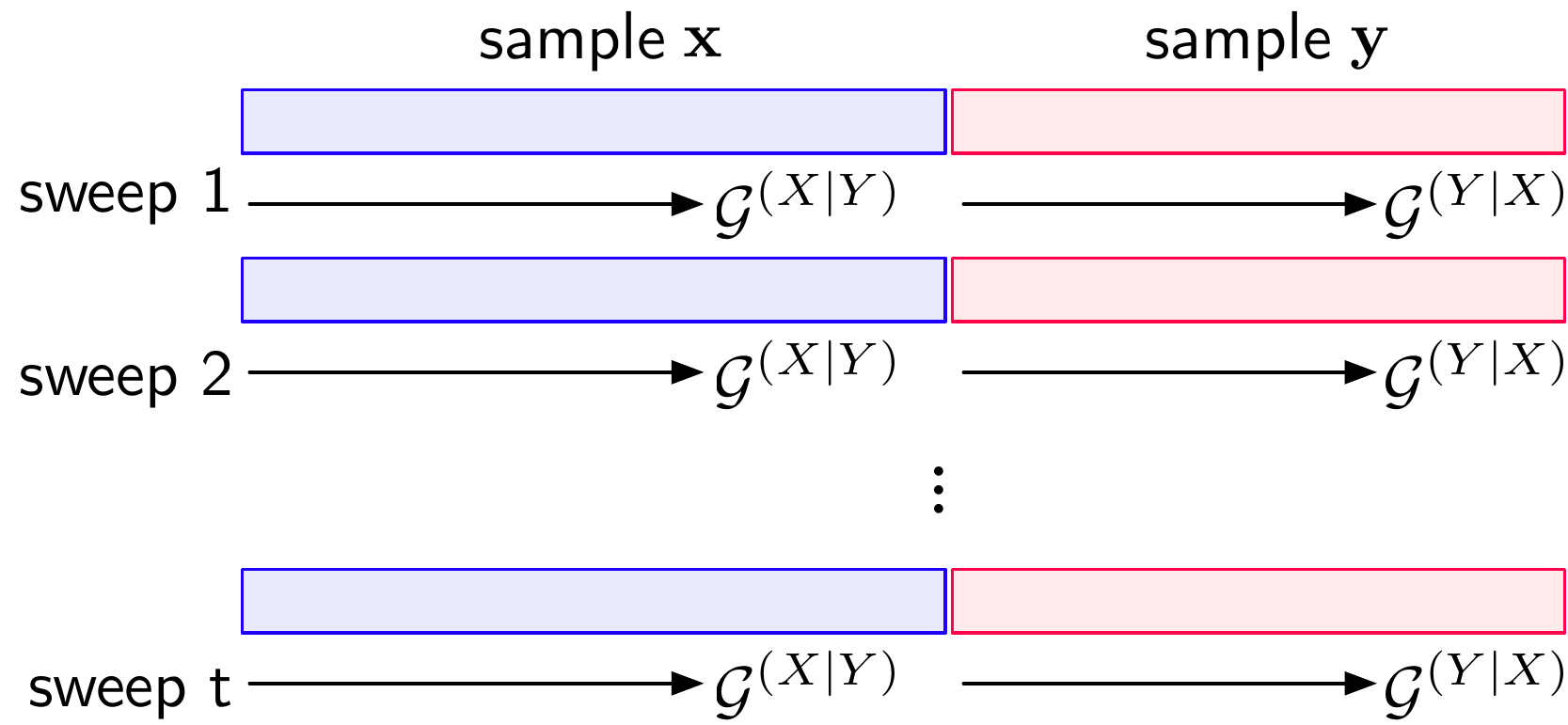}
			\caption{}
			\label{gibbscond}
		\end{subfigure}
		\caption{Sampling from a joint model $\mathcal{G}^{(X,Y)}$ (Figure~\ref{gibbsco})
			and two conditional models $\mathcal{G}^{(X|Y)}$ and $\mathcal{G}^{(Y|X)}$ (Figure~\ref{gibbscond})
			yields the same MAP estimates.
		}
		\label{fig:gibbs}
	\end{figure*}
	
	Continuing with the example from the previous section,
	in each iteration of Gibbs sampling (a ``sweep'' of samples),
	the sampler goes through each example in the data,
	which can be viewed as sampling from the full posterior of a joint model $\mathcal{G}^{(X,Y)}$ as in Figure~\ref{gibbsco}.
	Thus,
	when sampling an instance $x_i\in\mathbf{x}$,
	the collapsed conditional likelihood is
	\begin{align}
		\Pr\left(x = k | \mathbf{x}_-, \mathbf{y}; \alpha\right)
		~=~& \frac{\Pr(x = k, \mathbf{x}_-,\mathbf{y};\alpha)}{\Pr(\mathbf{x}_-,\mathbf{y};\alpha)} \\
		~=~& \frac{\Gamma\left(\alpha_k + n_{k|x} + n_{k|y}\right) }{\Gamma\left(N_x+N_y + \mathbf{1}^\top\alpha\right)} \cdot
		\frac{\Gamma\left(N_x+N_y^{(-i)} + \mathbf{1}^\top\alpha\right)}{\Gamma\left(\alpha_k + n_{k|x}^{(-i)} + n_{k|y}\right) }\\
		~=~&\frac{n_{k|x}^{(-i)}+n_{k|y}+\alpha_k}{N_x^{(-i)}+N_y+\mathbf{1}^\top \alpha},\label{gengibbs}
	\end{align}	where $\mathbf{x}_-$ is the set of tokens excluding the current one and $n_{k|x}^{(-i)}$ is the number of instances $x$ assigned to category $k$ except the current $x_i$.
	Note that in this equation,
	$\alpha$ is the hyperparameter for the Dirichlet prior,
	which gets added to the counts in the formula after integrating out the parameters $\theta$.

	Using our formulation from the previous section,
	we can separate each sweep into two \textit{subprocedures},
	one for each variable.
	When sampling an instance of $x_i\in\mathbf{x}$,
	the histogram of sample $\mathbf{y}$ is fixed,
	and therefore
	it is sampling from the conditional model of $\mathcal{G}^{(X|Y)}$.
	Thus,
	the conditional likelihood is
	\begin{align}
		\Pr\left(x = k | \mathbf{x}_-; \mathbf{y},\alpha,\mathcal{G}^{(X|Y)}\right)
		~=~& \frac{\Pr(x = k, \mathbf{x}_-;\mathbf{y},\alpha)}{\Pr(\mathbf{x}_-;\mathbf{y},\alpha)} \\
		~=~& \frac{\Gamma\left(n_{k|x} + ( n_{k|y} + \alpha_k)\right) }{\Gamma\left(N_x+(N_y + \mathbf{1}^\top\alpha)\right)} \cdot
		\frac{\Gamma\left(N_x+(N_y^{(-i)} + \mathbf{1}^\top\alpha)\right)}{\Gamma\left(n_{k|x}^{(-i)} + (n_{k|y} + \alpha_k)\right) }\\
		~=~&\frac{n_{k|x}^{(-i)}+(n_{k|y}+\alpha_k)}{N_x^{(-i)}+(N_y+\mathbf{1}^\top \alpha)},\label{condgibbs}
	\end{align}	
	where the hyperparameter for variable $X$ and category $k$ becomes $n_{k|y}+\alpha_k$.
	Similarly,
	when sampling $y_i\in\mathbf{y}$ which is generated from
	the model $\mathcal{G}^{(Y|X)}$,
	the conditional likelihood is
	\begin{align}
		\Pr\left(y = k | \mathbf{y}_-; \mathbf{x},\alpha,\mathcal{G}^{(Y|X)}\right)
		~=~&\frac{n_{k|y}^{(-i)}+(n_{k|x}+\alpha_k)}{N_y^{(-i)}+(N_x+\mathbf{1}^\top \alpha)},\label{condgibbs2}
	\end{align}	
	with $n_{k|x}+\alpha_k$ as the hyperparameter for $Y$.
	This process is shown in Figure~\ref{gibbscond}.

	From the calculation perspective,
	although the meaning of Equations~(\ref{gengibbs}), (\ref{condgibbs}) and (\ref{condgibbs2}) are different,
	their formulae are identical.
	This allows us to analyze similar models using the conditional formulation
	without changing the posterior estimation.
	A similar approach is the pseudo-likelihood approximation,
	where a joint model is reformulated as the combination of two conditional models,
	and the optimal parameters for the pseudo-likelihood function are the same as for the original joint likelihood function~\cite{besag1975statistical,pgmbook,LeppaahoPRC17}.

	\subsection{Transfer Operation}
	\label{sec:to}
	
	Now that we have made the transfer process explicit and showed that
	this alternative formulation yields same collapsed posterior,
	we are able to describe a similar process in detail in the context of multilingual topic models.

	If we treat $X$ and $Y$ in the previous example as two languages,
	and the samples $\mathbf{x}$ and $\mathbf{y}$ as either words, tokens, or documents from the two languages,
	we have a bilingual dataset $(\mathbf{x},\mathbf{y})$.
	Topic models have more complex graphical structures,
	where the examples (tokens) are
	organized within certain scopes, \textit{e.g.,} documents.
	To define the transfer process for a specific topic model,
	when generating samples in one language based on the transfer process of the model,
	we have to specify what examples we want to use from another language,
	how much, and where we want to use them.
	To this end,
	we define the \textbf{transfer operation},
	which allows us to examine different models under a unified framework to compare them systematically.

	\begin{definition}[Transfer operation]
		
		Let $\Omega\in\mathbb{R}^{M}$ be the target distribution of knowledge transfer with dimensionality $M$.
		A transfer operation on $\Omega$
		from language $\ell_1$ to $\ell_2$ is defined as a function
		\begin{align}
			h_{\Omega}:~\mathbb{R}^{L_2\times L_1}\times\mathbb{N}^{L_1\times M}\times\mathbb{R}_+^{M}~\mapsto~\mathbb{R}^{M},
		\end{align}
		where $L_1$ and $L_2$ are the relevant dimensionalities for languages $\ell_1$ and $\ell_2$.
	\end{definition}
	
	In this definition, the first argument of the transfer operation is where the two languages connect to each other,
	and can be defined as any bilingual supervision needed to enable transfer.
	The actual values of $L_1$ and $L_2$ depend on specific models.	
	For example, it could be an identity matrix ${\delta}$
	with $L_1$ and $L_2$ the number of documents in languages $\ell_1$ and $\ell_2$ respectively,
	where $\delta_{ij}=1$ if documents $i$ and $j$ are translations, or zero otherwise.
	This is the core of crosslingual transfer through the transfer operation;
	later we will see that different multilingual topic models mostly only differ in the input of this argument,
	and designing this matrix is critical for an efficient knowledge transfer.
	
	The second argument in the transfer operation is
	the sufficient statistics of the transfer source ($\ell_1$ in the definition).
	After generating instances in language $\ell_1$,
	the statistics are organized into a matrix.
	The last argument is a prior distribution over the possible target distributions $\Omega$.
	
\begin{figure}
		\centering
		\includegraphics[width=\linewidth]{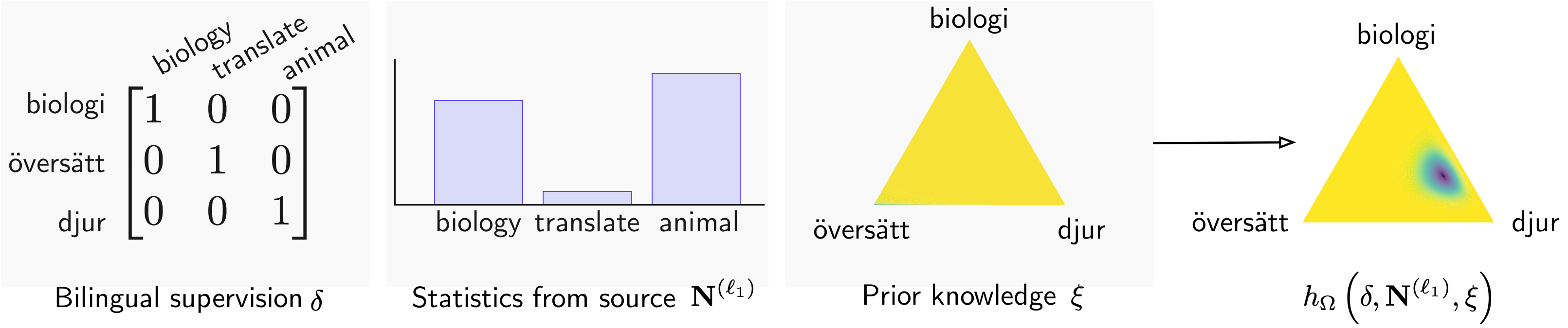}
		\caption{An illustration of a transfer operation on a 3-dimensional Dirichlet distribution.
			The first argument of $h_\Omega$ is a bilingual supervision $\delta$, which is a $3\times 3$ matrix
			indicating word translations between two languages.
			The second argument $\mathbf{N}^{(\ell_1)}$ is the sufficient statistics (or histogram)
			from the sample in language $\ell_1$,
			whose dimension is aligned with $\delta$.
			With $\xi$ as the prior knowledge (a symmetric hyperparameter),
			the result of $h_\Omega$ is then used as hyperparameters for the Dirichlet distribution.	
		}
		\label{fig:transferdir}
	\end{figure}

	The output of transfer operation depends on and has the same dimensionality as the target distribution,
	which will be used as the prior to generate a multinomial distribution.
	Let $\Omega$ be the target distribution from which a topic of language $\ell_2$ is generated,
	\textit{i.e.,} $z\sim\mathrm{Multinomial}(\Omega)$.
	With a transfer operation, 
	a topic is generated as follows:
	\begin{align*}
		\Omega &~\sim~ \mathrm{Dirichlet}\left(h_{\Omega}\left( {\delta}, \mathbf{N}^{(\ell_1)}, \xi \right)   \right),\\
		z &~\sim~ \mathrm{Multinomial}(\Omega),
	\end{align*}
	where ${\delta}$ is bilingual supervision, $\mathbf{N}^{(\ell_1)}$ the generated sample of language $\ell_1$,
	and ${\xi}$ a prior distribution with the same dimensionality as $\Omega$.
	See \fig{fig:transferdir} for an illustration.
	
	In summary, this definition highlights three elements that are necessary to enable transfer:
	\textbf{1)} language transformations or supervision from the transfer source to destination,
	\textbf{2)} data statistics in the source,
	and \textbf{3)} a prior on the destination.
	
	In the next section,
	we show how different topic models can be formulated with transfer operations, as well as how transfer operations can be used in the design of new models.

	\section{Representative Models}
	\label{sec:models}
	
\begin{figure}
		\centering
		\includegraphics[width=\linewidth]{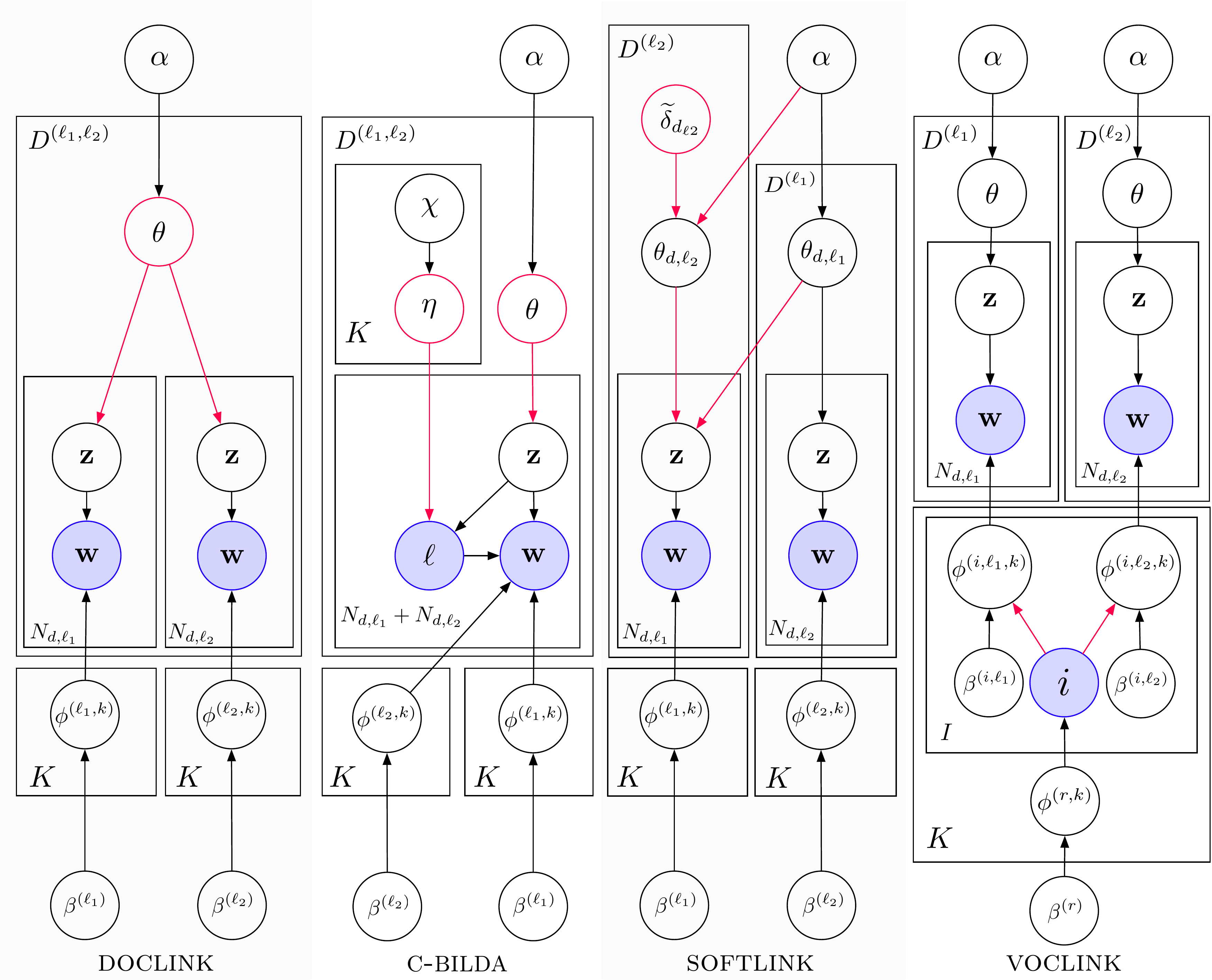}
		\caption{Plate notations of \doclink{}, \cbilda{}, \softlink{}, and \voclink{} (from left to right). We use red lines to make the knowledge transfer component clear. Note that in \voclink{} we assume every word is translated, so the plate notation does not include untranslated words.}
		\label{fig:platesss}
	\end{figure}

	In this section,
	we describe four representative multilingual topic models
	in terms of the transfer operation formulation.
	These are also the models we will experiment on in Section~\ref{sec:expr}.
	The plate notations of these models are shown in Figure~\ref{fig:platesss},
	and we provide notations frequently used in these models in Table~\ref{nt}.
	
	\begin{table}[h]
		\centering
		\begin{tabular}{l|p{0.8\linewidth}}
			\hline 
			{Notations} & {Descriptions} \\  \hline \hline
			$z$ & The topic assignment to a token. \\ 
			\hl $w^{(\ell)}$ &\hl  A word type in language $\ell$. \\ 
			$V^{(\ell)}$ & The size of vocabulary in language $\ell$.\\ 
			\hl $D^{(\ell)}$ & \hl The size of corpus in language $\ell$. \\ 
			$D^{(\ell_1,\ell_2)}$ &  The number of document pairs in languages $\ell_1$ and $\ell_2$. \\ 
			\hl $\alpha$ & \hl A symmetric Dirichlet prior vector of size $K$, where $K$ is the number of topics, and each cell is denoted as $\alpha_k$.\\ 
			$\theta_{d,\ell}$ &  Multinomial distribution over topics for a document $d$ in language $\ell$. \\  
			\hl $\beta^{(\ell)}$ & \hl  A symmetric Dirichlet prior vector of size $V^{(\ell)}$, where $V^{(\ell)}$ is the size of vocabulary in language $\ell$. \\ 
			$\beta^{(r,\ell)}$ &  An asymmetric Dirichlet prior vector of size $I+V^{(\ell,-)}$, where $I$ is the number of internal nodes in a Dirichlet tree, and $V^{(\ell,-)}$ the number of untranslated words in language $\ell$. Each cell is denoted as $\beta^{(r,\ell)}_i$, indicating a scalar prior to a specific node $i$ or an untranslated word type. \\ 
			\hl $\beta^{(i,\ell)}$ & \hl A symmetric Dirichlet prior vector of size $V_i^{(\ell)}$, where $V_i^{(\ell)}$ is the number of word types in language $\ell$ under internal node $i$. \\ 
			$\phi^{(\ell,k)}$ &  Multinomial distribution over word types in language $\ell$ of topic $k$ for topic $k$. \\ 	
			\hl $\phi^{(r,\ell,k)}$ &\hl  Multinomial distribution over internal nodes in a Dirichlet tree for topic $k$. \\ 
			$\phi^{(i,\ell,k)}$ &  Multinomial distribution over all word types in language $\ell$ under internal node $i$ for topic $k$. \\ \hline
		\end{tabular}
		\caption{Notation table.}
		\label{nt}
	\end{table}

	\subsection{Standard Models}
	
	Typical multilingual topic models are designed based on
	simple observations of multilingual data,
	such as parallel corpora and dictionaries.
	We focus on three popular models,
	and re-formulate them using the conditional generation assumption
	and the transfer operation we introduced in the previous sections.

	\subsubsection{\doclink{}}\label{ex:doclink}

	The document links model (\doclink{}) uses parallel/comparable datasets,
	so that each bilingual document pair shares the same distribution over topics.
	Assume the document $d$ in language $\ell_1$ is paired with $d$ in language $\ell_2$.
	Thus, the transfer target distribution is $\theta_{d,\ell_2}\in\mathbb{R}^K$
	where $K$ is the number of topics.
	For a document $d_{\ell 2}$,
	let $\delta\in\mathbb{N}_+^{D^{(\ell_1)}}$ be an indicator vector to indicate if a document $d_{\ell 1}$
	is a translation or comparable document to $d_{\ell 2}$,
	\begin{align}
		\delta_{d_{\ell 1}} ~=~ \mathds{1}\big\{\ d_{\ell 2} \text{\ and\ } d_{\ell 1} \text{\ are translations}\ \big\},\label{ident}
	\end{align}
	where $D^{(\ell_1)}$ is the number of documents in language $\ell_1$.
	Thus, the transfer operation for each document $d_{\ell 2}$ can be defined as
	\begin{align}
		h_{\theta_{d,\ell_2}}\left(\delta,\mathbf{N}^{(\ell_1)},\alpha\right) ~=~ \delta\cdot\vt{N}^{(\ell_1)} + \alpha,\label{doclinktransfer}
	\end{align} where
	$\mathbf{N}^{(\ell_1)}\in\mathbb{N}^{D^{\left(\ell_1\right)}\times K}$
	is the sufficient statistics from language $\ell_1$,
	and each cell $n_{dk}$ is the count of topic $k$ appearing in document $d$.
	We call this a ``document level'' model,
	since the transfer target distribution is document-wise.
	
	On the other hand, \doclink{} does not any word-level knowledge, such as dictionaries, so
	the transfer operation on $\phi$ in \doclink{} is straightforward.
	For every topic $k=1,\ldots,K$ and each word type $w$ regardless of its language,
	\begin{align}
		h_{\phi^{(\ell_2,k)}}\left(\mathbf{0},\mathbf{N}^{(\ell_1)},\beta^{(\ell_2)}\right)~=~\mathbf{0}\cdot\mathbf{N}^{(\ell_1)}+\beta^{(\ell_2)}~=~\beta^{(\ell_2)},\label{doclinktransfer2}
	\end{align}
	where $\beta^{(\ell_2)}\in\mathbb{R}^{V^{(\ell_2)}}_+$ is a symmetric Dirichlet prior for the topic-vocabulary distributions $\phi^{(\ell_2,k)}$, and $V^{(\ell_2)}$ is the size of vocabulary in language $\ell_2$.

	\subsubsection{\cbilda{}}\label{ex:cbi}
	As a variation of \doclink{},
	\cbilda{} has all of the components of \doclink{}
	and has the same transfer operations on $\theta$ and $\phi$ as in Equations~(\ref{doclinktransfer}) and (\ref{doclinktransfer2}),
	so this model is considered as a document-level model as well.
	Recall that \cbilda{} additionally models topic-language distributions $\eta$.\footnote{
		The original notation for topic-language distribution is $\delta$~\cite{HeymanVM16}.
		To avoid confusion in \eq{ident}, we change to $\eta$.
		We also follow the original paper where the model is for a bilingual case.
	}
	For each document pair $d$ and each topic $k$,
	a bivariate Bernoulli distribution over the two languages $\eta^{(k,{d})}\in\mathbb{R}_+^2$ is drawn from a Beta distribution parameterized by $\left(\chi^{(d,\ell_1)},\chi^{(d,\ell_2)}\right)$:
	\begin{align}
		\eta^{(k,{d})} ~\sim~ & \mathrm{Beta}\left(\chi^{(d,\ell_1)},\chi^{(d,\ell_2)}\right), \\
		\ell^{(k,m)} ~\sim~ & \mathrm{Bernoulli}\left(\eta^{(k,{d})}\right),
	\end{align}
	where $\ell^{(k,m)}$ is the language of the $m$-th token assigned to topic $k$ in the \textit{entire} document pair $d$.
	Intuitively, 
	$\eta^{(k,{d})}_\ell$ is the probability of generating a token in language $\ell$
	given the current document pair $d$ and topic $k$.
	
\begin{figure}
		\centering
		\includegraphics[width=\linewidth]{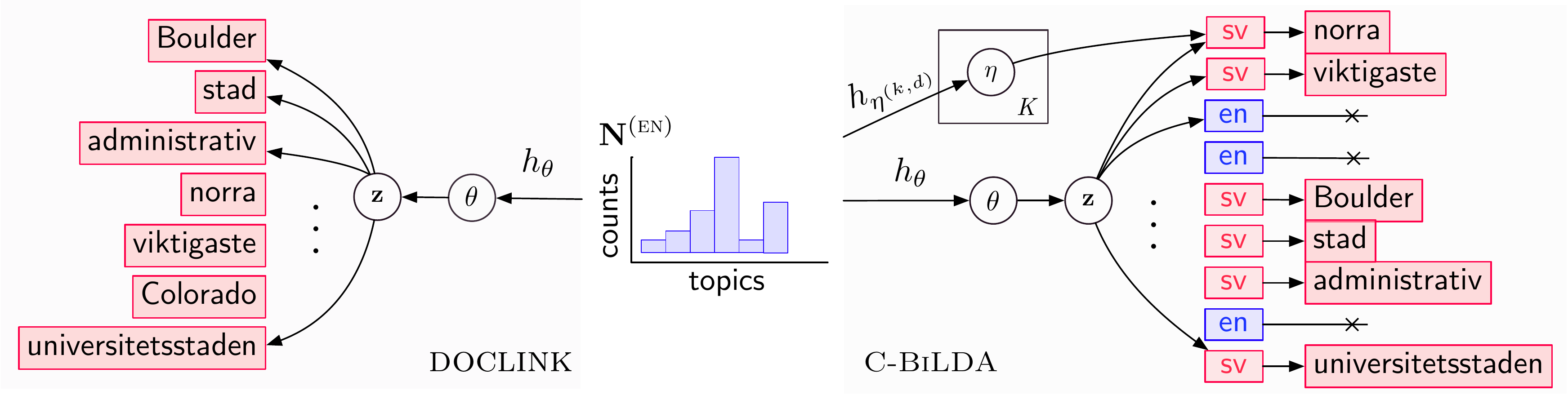}
		\caption{An illustration of difference between \doclink{} and \cbilda{} in sequential generating process. \doclink{} uses transfer operation on $\theta$ to generate topics and then word types in Swedish (\sv{}). Additionally, \cbilda{} uses a transfer operation on $\eta$ to generate a language label according to a topic $z$. If the language generated is in Swedish, it draws a word type from the vocabulary; otherwise, the token is discarded.}
		\label{fig:selector}
	\end{figure}

	Before diving into the specific definition of the transfer operation for this model,
	we need to take a closer look at the generative process of \cbilda{} first,
	since in this model, language itself is a random variable as well.
	We describe the generative process in terms of the conditional formulation where one language is conditioned on the other.
	As usual, a monolingual model first generates documents in $\ell_1$,
	and at this point each document pair $d$ only has tokens in one language.
	Then for each document pair $d$,
	the conditional model additionally generates a number of topics $z$
	using the transfer operation on $\theta$ as defined in Equation~(\ref{doclinktransfer}).
	Instead of directly drawing a new word type in language $\ell_2$ according to $z$,
	\cbilda{} adds a step to generate a language $\ell'$ from $\eta^{(z,d)}$.
	Since the current token is supposed to be in language $\ell_2$,
	if $\ell' \neq \ell_2$, this token is dropped, and the model keeps drawing the next topic $z$;
	otherwise, a word type is drawn from $\phi^{(z,\ell_2)}$
	and attached to the document pair $d$.
	Once this process is over, each document pair $d$ contains tokens from two languages,
	and by separating the tokens based on their languages
	we can obtain the corresponding set of comparable document pairs.
	Conceptually, \cbilda{} adds an additional ``selector'' in the generative process,
	to decide if a topic should appear more in $\ell_2$ based on topics in $\ell_1$.
	We use Figure~\ref{fig:selector} as an illustration to show the difference between \doclink{} and \cbilda{}.
	
	It is clear that the generation of tokens in language $\ell_2$ is affected
	by that of language $\ell_1$,
	thus we define an additional transfer operation on $\eta^{(k,d)}$.
	The bilingual supervision $\delta$ is the same as Equation~(\ref{ident}),
	which is a vector of dimension $D^{(\ell_1)}$ indicating document translations.
	We denote the statistics term $\mathbf{N}^{(\ell_1)}_k\in\mathbb{R}^{D^{(\ell_1)}\times 2}$,
	where each cell in the first column $n_{dk}$ is the counts of topic $k$ in document $d$,
	while the second column is a zero vector.
	Lastly, the prior term is also a two-dimensional vector $\chi^{(d)}=\left( \chi^{(d,\ell_1)},\chi^{(d,\ell_1)} \right)$.
	Together, we have the transfer operation defined as
	\begin{align}
		h_{\eta^{(k,d)}}\left(\delta, \mathbf{N}^{(\ell_1)}_k, \chi^{(d)}\right)
		~=~ \delta\cdot \mathbf{N}^{(\ell_1)}_k + \chi^{(d)}.
	\end{align}

	\subsubsection{\voclink{}}\label{ex:voclink}
	\namecite{JagarlamudiD10} and \namecite{BoydGraberB09} introduced
	another type of multilingual topic model which uses a dictionary for word-level supervision instead of parallel/comparable
	documents as supervision,
	and we call this model \voclink{}.\footnote{
		While some models as in~\namecite{HuZEB14} transfers knowledge at both document and word levels,
		in this analysis, we only focus on the word level where no transfer happens on document level.
		The generalization simply involves using the same transfer operation on $\theta$ that is used in \doclink{}.
	}
	Since no document-level supervision is used,
	the transfer operation on $\theta$ is simply defined as
	\begin{align}
		h_{\theta_{d,\ell_2}}\left(\mathbf{0},\mathbf{N}^{(\ell_1)},\alpha\right) ~=~ \mathbf{0}\cdot\vt{N}^{(\ell_1)} + \alpha ~=~ \alpha.
	\end{align}
	
	We now construct the transfer operation on the topic-word distribution $\phi$ based on
	the tree-structued priors in \voclink{} (\fig{tree}).	
	Recall that each word $w^{(\ell)}$ is associated with at least one path, denoted as $\lambda_{w^{(\ell)}}$.
	If $w^{(\ell)}$ is translated,
	the path is $\lambda_{w^{(\ell)}} = \left(r\rightarrow i, i \rightarrow w^{(\ell)}\right)$
	where $r$ is the root and $i$ an internal node;
	otherwise, the path is simply the edge from root to that word.
	Thus, on the first level of the tree,
	the Dirichlet distribution $\phi^{(r,\ell_2,k)}$ is of dimension $I + V^{(\ell_2,-)}$,
	where $I$ is the number of internal nodes (\textit{i.e.,} word translation entries),
	and $V^{(\ell_2,-)}$ the untranslated word types in language $\ell_2$.
	Let $\delta\in\mathbb{R}_+^{\left(I + V^{(\ell_2,-)}\right)\times V_1}$ be
	an indicator matrix where $V_1$ is the number of translated words in language $\ell_1$,
	and	each cell is
	\begin{align}
		\delta_{i,w^{(\ell_1)}} ~=~ \mathds{1} \left\{\ w^{(\ell_1)} \text{\ is under node\ } i\ \right\}.
	\end{align}
	Given a topic $k$,
	the statistics argument $\mathbf{N}^{(\ell_1)}\in\mathbb{R}^{V_1}$
	is a vector where each cell $n_{w}$ is the count of word $w$ assigned to topic $k$.
	Note that in the tree structure,
	the prior for Dirichlet is asymmetric and is scaled by the number of translations under each internal node.
	Thus, the transfer operation on $\phi^{(r,\ell_2,k)}$ is
	\begin{align}
		h_{\phi^{(r,\ell_2,k)}}\left( \delta, \mathbf{N}^{(\ell_1)}, \beta^{(r,\ell_2)} \right)
		~=~ \delta\cdot \mathbf{N}^{(\ell_1)} + \beta^{(r,\ell_2)}.\label{voctr}
	\end{align}
	Under each internal node, the Dirichlet is only related to specific languages,
	so no transfer happens,
	and the transfer operation on $\phi^{(i,\ell_2,k)}$ for an internal node $i$
	is simply $\beta^{(i,\ell_2)}$:
	\begin{align}
		h_{\phi^{(i,\ell_2,k)}}\left( \mathbf{0}, \mathbf{N}^{(\ell_1)}, \beta^{(i,\ell_2)} \right)
		~=~ \mathbf{0}\cdot \mathbf{N}^{(\ell_1)} + \beta^{(i,\ell_2)} = \beta^{(i,\ell_2)}.
	\end{align}

	\subsection{\softlink{}: A Transfer Operation-Based Model}\label{ex:softlink}
	
	We have formulated three representative multilingual topic models
	by defining transfer operations for each model above.
	Our recent work, called \softlink{}~\cite{HaoP18},
	is explicitly designed according to the understanding of this transfer process.
	We present this model
	as a demonstration of how transfer operations can be used to build new multilingual topic models, which might not have an equivalent  formulation using the standard co-generation model, by modifying the transfer operation.
	
	In \doclink{}, the supervision argument $\delta$ in the transfer operation
	is constructed using comparable datasets.
	This requirement, however, substantially limits the data that can be used.
	Moreover, the supervision $\delta$ is also limited by the data;
	if there's no translation available to a target document, $\delta$ is an all-zero vector,
	and the transfer operation defined in Equation~(\ref{doclinktransfer})
	will cancel out all the available information $\mathbf{N}^{(\ell_1)}$ for the target document,
	which is an ineffective use of the resource.
	Unlike parallel corpora,
	dictionaries are widely available and often easy to obtain for many languages.
	Thus, the general idea of \softlink{} is to use a dictionary to 
	retrieve as much as possible information from $\ell_1$ to construct $\delta$ in a way that links potentially comparable documents together,
	even if the corpus itself does not explicitly link together documents.
	
\begin{figure}
		\centering
		\includegraphics[width=\linewidth]{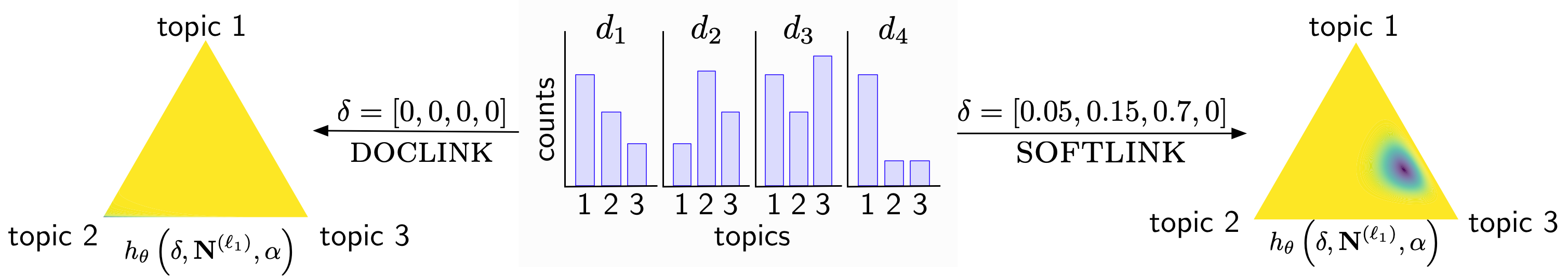}
		\caption{An example of how different inputs of transfer operation result in
			different Dirichlet priors through \doclink{} and \softlink{}.
			The middle is a mini-corpus in language $\ell_1$ and each document's topic histogram.
			When a document in $\ell_2$ is not translation to any of those in $\ell_1$,
			\doclink{} defines $\delta$ as an all-zero vector which leads to an uninformative
			symmetric prior.
			In contrast, \softlink{} uses a dictionary to create $\delta$ as a distribution
			so that the topic histogram in each document in $\ell_1$
			can still be proportionally transferred.}
		\label{fig:transfer-dir}
	\end{figure}

	Specifically, for a document $d_{\ell 2}$, instead of a pre-defined indicator vector,
	\softlink{} defines $\delta$ as a probabilistic distribution over all documents in language $\ell_1$:
	\begin{align}
		\delta_{d_{\ell 1}} ~\propto~\frac{|\left\{w^{(\ell_1)}\right\}\cap\left\{w^{(\ell_2)}\right\}|}{|\left\{w^{(\ell_1)}\right\}\cup\left\{w^{(\ell_2)}\right\}|},\label{eq:softdelta}
	\end{align}
	where $\{w^{(\ell)}\}$ contains all the word types that appear in document $d_{\ell}$,
	and $\left\{w^{(\ell_1)}\right\}\cap\left\{w^{(\ell_2)}\right\}$ indicates all word pairs $\left(w^{(\ell_1)},w^{(\ell_2)}\right)$ in a dictionary as translations.
	Thus, $\delta_{d_{\ell 1}}$ can be interpreted 
	as the ``probability'' of $d_{\ell 1}$ being translation to $d_{\ell 2}$.
	We call $\delta$ the ``transfer distribution''.
	See Figure~\ref{fig:transfer-dir} for an illustration.

	In our initial work, we show that instead of a dense distribution,
	it is more efficient to make the transfer distributions sparse by thresholding,
	\begin{align}
		\widetilde{\delta}_{d_{\ell 1}} ~\propto~ \mathds{1}\big\{\delta_{d_{\ell 1}} > \pi \cdot \max\left(\delta\right)\big\}\cdot\delta_{d_{\ell 1}},
	\end{align}
	where $\pi\in[0,1]$ is a fixed threshold parameter.
	With the same definition of $\mathbf{N}^{(\ell_1)}$ and $\alpha$ in Equation~(\ref{doclinktransfer})
	and $\delta$ defined as Equation~(\ref{eq:softdelta}),
	\softlink{} completes the same transfer operations,
	\begin{align}
		h_{\theta_{d,\ell_2}}\left(\delta,\mathbf{N}^{(\ell_1)},\alpha\right) &~=~ \delta\cdot\vt{N}^{(\ell_1)} + \alpha,\\
		h_{\phi^{(\ell_2,k)}}\left(\mathbf{0},\mathbf{N}^{(\ell_1)},\beta^{(\ell_2)}\right)&~=~\mathbf{0}\cdot\mathbf{N}^{(\ell_1)}+\beta^{(\ell_2)}~=~\beta^{(\ell_2)}.
	\end{align}

	\begin{table}
		\centering
		\begin{tabular}{c|l|l|p{0.4\textwidth}}
			\hline
			Model & Document level & Word level & Parameters of $h$  \\ \hline\hline
			\hl LDA & \hl $\alpha$ & \hl $\beta^{(\ell_2)}$ & \hl --- \\ 
			\doclink{} & $\delta\cdot\mathbf{N}^{(\ell_1)}+\alpha$ & $\beta^{(\ell_2)}$ &
			\multirow{3}{*}{
				\hspace{-0.5em}\makecell[l]{
					$\delta$: indicator vector;\\
					$\vt{N}^{(\ell_1)}$: doc-by-topic matrix;\\
					supervision: comparable documents;\\
			}} \\ \cline{1-3} 
			\multirow{2}{*}{\cbilda{}} & $\delta\cdot\mathbf{N}^{(\ell_1)}+\alpha,$ & \multirow{2}{*}{$\beta^{(\ell_2)}$} & \\
			& $\delta\cdot\mathbf{N}^{(\ell_1)}_k+\chi^{(d)}$ & &\\ 
			\hl \softlink{} & \hl $\delta\cdot\mathbf{N}^{(\ell_1)}+\alpha$ & \hl $\beta^{(\ell_2)}$ &
			\hl \makecell[l]{
				$\delta$: transfer distribution;\\
				$\vt{N}^{(\ell_1)}$: doc-by-topic matrix;\\
				supervision: dictionary;\\
			}
			\\  
			\voclink{} & $\alpha$ & $\delta\cdot \mathbf{N}^{(\ell_1)} + \beta^{(r,\ell_2)}$ &
			\makecell[l]{
				$\delta$: indicator vector;\\
				$\vt{N}^{(\ell_1)}$: node-by-word matrix;\\
				supervision: dictionary;\\
			}\\ \hline
		\end{tabular}
		\caption{Summary of transfer operations defined in the compared models, where we assume the direction of transfer is from $\ell_1$ to $\ell_2$.}
		\label{tab:sum}
	\end{table}
	
	\subsection{Summary: Transfer Levels and Transfer Models}
	
	We categorize transfer operations into two groups
	based on the target transfer distribution.
	\textit{Document-level} operations transfer knowledge on distributions related to the entire document,
	such as $\theta$ in \doclink{}, \cbilda{} and \softlink{}, and $\eta$ in \cbilda{}.
	\textit{Word-level} operations transfer knowledge on those related to the entire vocabulary or specific word types,
	such as $\phi$ in \voclink{}.
	
	When a model only has transfer operations on just one specific level,
	we also use the transfer level to refer the model.
	For example, \doclink{}, \cbilda{} and \softlink{} are all document-level models,
	while \voclink{} is a word-level model.
	Those that transfer knowledge on multiple levels such as~\namecite{HuZEB14}
	are called mixed-level models.
	
	We summarize the transfer operation definitions for different models in Table~\ref{tab:sum},
	and add monolingual LDA as a reference to show how transfer operations are defined when no transfer takes place.
	We will experiment on the four multilingual models in Sections~\ref{ex:doclink} through \ref{ex:softlink}.

	\section{Experiment Settings}
	\label{sec:expr}

	From discussions above,
	we are able to describe various multilingual topic models
	by defining different transfer operations,
	which explicitly represent the language transfer process.
	When designing and applying those transfer operations in practice,
	some natural questions arise, such as
	which transfer operation is more effective in what kind of situations,
	and
	how to design a model that is more generalizable regardless of availability of multilingual resources.
	
	To study the model behaviors empirically,
	we train the four models described in the previous section, \textit{i.e.,} \doclink{},
	\cbilda{}, \softlink{}, and \voclink{}, in ten languages.
	Considering the resources available,
	we separate the ten languages into two groups:
	high-resource languages (\highlan{}) and low-resource languages (\lowlan{}).
	For \highlan{},
	we have relatively abundant resources such as dictionary entries and document translations.
	We additionally use these languages to simulate the settings of \lowlan{}
	by training multilingual topic models with different amount of resources.
	For \lowlan{}, we use all resources available to verify experiment results
	and conclusions from \highlan{}.

	\subsection{Language Groups and Preprocessing}
	
	We separate the ten languages into two groups---\highlan{} and \lowlan{}.
	In this section, we describe the preprocessing details
	of these languages.
	
	\subsubsection{\highlan{}}
	Languages in this group have a relatively large amount of resources,
	and have been widely experimented on in multilingual studies.
	Considering language diversity,
	we select representative languages from five different families:
	Arabic (\ar{}, Semitic),
	German (\de{}, Germanic),
	Spanish (\es{}, Romance),
	Russian (\ru{}, Slavic),
	and Chinese (\zh{}, Sinitic).
	We follow standard preprocessing procedures:
	we first use stemmers to process both documents and dictionaries (segmenter for Chinese),
	then we remove stopwords based on a fixed list
	and the most $100$ frequent word types in the training corpus.
	The tools for preprocessing are listed in Table~\ref{tools}.
	\begin{table}
		\centering
		\begin{tabular}{c|c|c|c}
			\hline 
			Language & Family & Stemmer & Stopwords \\ \hline \hline
			\en{} & Germanic & \texttt{SnowBallStemmer}~\footnotemark &  NLTK \\ \hline 
			\de{} & Germanic & \texttt{SnowBallStemmer} &  NLTK \\ \hline 
			\es{} & Romance & \texttt{SnowBallStemmer} & NLTK  \\ \hline 
			\ru{} & Slavic & \texttt{SnowBallStemmer} &  NLTK \\ \hline 
			\ar{} & Semitic & \texttt{Assem's Arabic Light Stemmer}~\footnotemark &  GitHub~\footnotemark \\ \hline
			\zh{} & Sinitic & \texttt{Jieba}~\footnotemark &  GitHub \\ \hline 
		\end{tabular}
		\caption{List of source of stemmers and stopwords used in experiments for \highlan{}.}
		\label{tools}
	\end{table}
	
	\footnotetext[4]{\url{http://snowball.tartarus.org};}
	\footnotetext[5]{\url{http://arabicstemmer.com};}
	\footnotetext[6]{\url{https://github.com/6/stopwords-json};}
	\footnotetext[7]{\url{https://github.com/fxsjy/jieba}.}
	
	\subsubsection{\lowlan{}}	 
	Languages in this group have much fewer resources than those in \highlan{},
	considered as low-resource languages.
	We similarly select five languages from different families:
	Amharic (\am{}, Afro-Asiatic),
	Aymara (\ay{}, Aymaran),
	Macedonian (\mk{}, Indo-European),
	Swahili (\sw{}, Niger-Congo),
	and Tagalog (\tl{}, Austronesian).
	Note that some of these are not strictly ``low-resource''
	compared to many endangered languages.
	For the truly low-resource languages,
	it is very difficult to test the models with enough data,
	and therefore,
	we choose languages that are understudied in natural language processing literature.
	
	Preprocessing in this language group needs more consideration.
	Because they represent low-resource languages 
	that most natural language processing tools are not available for,
	we do not use a fixed stopword list.
	Stemmers are also not available for these languages,
	so we do not apply stemming.

	\subsection{Training Sets and Model Configurations}
	There are many resources available for multilingual research,
	such as the European Parliament Proceedings parallel corpus (\textsc{EuroParl},~\namecite{europarl}),
	The Bible, and Wikipedia.
	\text{EuroParl} provides a perfectly parallel corpus with precise translations,
	but it only contains $21$ European languages,
	which limits its generalizability to most of the languages.
	The Bible, on the other hand, is also perfectly parallel
	and is widely available in $2{,}530$ languages.\footnote{
		Reported by the United Bible Societies: \url{https://www.unitedbiblesocieties.org/}}
	Its disadvantages, however, are that
	the contents are very limited (mostly about family and religion),
	the dataset size is small ($1{,}189$ chapters),
	and many languages do not have digital format~\cite{Christodoulopoulos15}.
	
	Compared to \textsc{EuroParl} and The Bible,
	Wikipedia provides comparable documents in many languages
	with a large range of content,
	making it a very popular choice for many multilingual studies.
	In our experiments,
	we create ten bilingual Wikipedia corpora,
	each containing documents in one of the languages in either \highlan{} or \lowlan{},
	paired with documents in English (\en{}).
	Though most multilingual topic models are not restricted to training bilingual corpora paired with English,
	this is a helpful way to focus our experiments and analysis.
	
	We present the statistics of the training corpus of Wikipedia
	and the dictionary we use (from Wiktionary) in the experiments in Table~\ref{wiki}.
	Note that we train topic models on bilingual pairs,
	where one of the languages is always English,
	so in the table we show statistics of English in every bilingual pair as well.
	
	\begin{table}\centering
		\resizebox{\textwidth}{!}{
			\begin{tabular}{c|c|c|r|r|c|r|r|c} \hline
				& & \multicolumn{3}{c|}{ English (\en{}) } & \multicolumn{3}{c|}{ Paired language } & Wiktionary \\ \hline
				& & \#docs & \#tokens & \#types & \#docs & \#tokens & \#types & \#entries\\ \hline \hline
				\parbox[t]{2mm}{\multirow{5}{*}{\rotatebox[origin=c]{90}{\highlan}}}
				& \ar{} & 2{,}000 & \phantom{0,}616{,}524 & 48{,}133  & 2{,}000 &  181{,}946 & 25{,}510 & 16{,}127\\ \cline{2-9}
				& \de{} & 2{,}000 & \phantom{0,}332{,}794 & 35{,}921  & 2{,}000 &  254{,}179 & 55{,}610 & 32{,}225 \\ \cline{2-9}
				& \es{} & 2{,}000 & \phantom{0,}369{,}181 & 37{,}100 &  2{,}000 &  239{,}189 & 30{,}258 & 31{,}563\\ \cline{2-9}
				& \ru{} & 2{,}000 & \phantom{0,}410{,}530 & 39{,}870 &  2{,}000 &  227{,}987 & 37{,}928 & 33{,}574\\ \cline{2-9}
				& \zh{} & 2{,}000 & \phantom{0,}392{,}745 & 38{,}217 &  2{,}000 &  168{,}804 & 44{,}228 & 23{,}276 \\ \hline
				\parbox[t]{2mm}{\multirow{5}{*}{\rotatebox[origin=c]{90}{\lowlan}}}
				& \am{} & 2,000 & 3,589,268 & 161,879 & 2,000  & 251,708 & 65,368 & \phantom{0}4{,}588 \\ \cline{2-9}
				& \ay{} & 2,000 & 1,758,811 & 84,064 & 2,000 & 169,439 & 24,136 & \phantom{0}1{,}982 \\ \cline{2-9}
				& \mk{} & 2,000 & 1,777,081 & 100,767  & 2,000 & 489,953 & 87,329  & \phantom{0}6,895 \\ \cline{2-9}
				& \sw{} & 2,000 & 2,513,838  & 143,691 & 2,000 & 353,038 & 46,359 & 15{,}257\\ \cline{2-9}
				& \tl{} & 2,000 & 2,017,643 & 261,919 & 2,000 & 232,891 & 41,618  & \phantom{0}6,552 \\ \hline
			\end{tabular}
		}
		\caption{Statistics of training Wikipedia corpus and Wiktionary.}
		\label{wiki}
	\end{table}
	
	Lastly,
	we summarize the model configurations in Table~\ref{modelset}.
	The goal of this study is to bring current multilingual topic models together,
	studying their corresponding strengths and limitations.
	To keep the experiments as comparable as possible,
	we use constant hyperparameters that are consistent across the models.
	For all models, we set the Dirichlet hyperparameter $\alpha_k=0.1$ for each topic $k=1,\ldots,K$.
	We run $1{,}000$ Gibbs sampling iterations on the training set
	and $200$ iterations on the test sets.
	The number of topics $K$ is set to $20$ by default for efficiency reasons.

	\begin{table}
		\centering
		\begin{tabular}{c|p{0.8\linewidth}}
			\hline 
			{Model} & {Hyperparameters} \\  \hline \hline 
			{\doclink{}} & We set $\beta$ to be a symmetric vector where each cell $\beta_i=0.01$ for all word types of all the languages,
			and use the \texttt{MALLET} implementation for training~\cite{McCallumMALLET}. To enable consistent comparison,
			we disable hyperparameter optimization provided in the package.\\  \hline 
			{\cbilda{}} & Following the experiment results from \namecite{HeymanVM16}, we set $\chi=2$ to make the results more competitive to \doclink{}. The rest of settings is same to \doclink{}.\\  \hline 
			{\softlink{}} & We use the document-wise thresholding approach for calculating the transfer distributions. The focus threshold is set to $0.8$. The rest of settings is same to \doclink{}. \\  \hline 
			{\voclink{}} & We set the scalar $\beta'=0.01$ for hyperparameter $\beta^{(r,\ell)}$ from the root to both internal nodes or leaves.
			For those from internal nodes to leaves, we set $\beta''=100$, following the settings in \namecite{HuZEB14}.  \\  \hline
		\end{tabular}
		\caption{Model specifications.}
		\label{modelset}
	\end{table}

	\subsection{Evaluation}

	We evaluate all models using both intrinsic and extrinsic metrics.
	Intrinsic evaluation is used to measure
	the topic quality or coherence learned from the training set,
	while extrinsic evaluation measures performance after applying the trained distributions to downstream crosslingual applications.
	For all the following experiments and tasks,
	we start by analyzing languages in \highlan{}.
	Then we apply the analyzed results to \lowlan{}.

	We choose topic coherence~\cite{HaoBGP18} and crosslingual document classification~\cite{SmetTM11} as intrinsic
	and extrinsic evaluation tasks, respectively.
	The reason for choosing these two tasks
	is that they examine the models from different angles:
	topic coherence looks at topic-word distributions,
	while classification focuses on document-topic distributions.
	Other evaluation tasks, 
	such as word translation detection and crosslingual information retrieval,
	also utilize the trained distributions,
	but here we focus on a straightforward and representative task.

	\subsubsection{Intrinsic Evaluation: Topic Quality}
	\label{sec:in}
	
	Intrinsic evaluation refers to evaluating the learned model directly without applying it to any particular task; for topic models, this is sually based on the quality of the topics.
	Standard evaluation measures for monolingual models,
	such as perplexity (or held-out likelihood, \namecite{WallachMSM09}) and
	Normalized Pointwise Mutual Information~(\npmi{}, \namecite{LauNB14}),
	could potentially be considered for crosslingual models.
	However, when evaluating multilingual topics,
	how words in different languages make sense together is also a critical criteria
	in addition to coherence within each of the languages.
	
	In monolingual studies, \namecite{ChangBGWB09} show that held-out likelihood
	is not always positively correlated with human judgments of topics.
	Held-out likelihood is additionally suboptimal for multilingual topic models,
	since this measure is only calculated \textit{within} each language,
	and the important crosslingual information is ignored.
	
	Crosslingual Normalized Pointwise Mutual Information~(\cnpmi{}, \namecite{HaoBGP18})
	is a measure designed specifically for multilingual topic models.
	Extended from the widely-used \npmi{} to measure topic quality in multilingual settings,
	\cnpmi{} uses a parallel reference corpus to extract crosslingual coherence.
	\cnpmi{} correlates well with bilingual speakers' judgments on topic quality and
	predictive performance in downstream applications.
	Therefore, we use \cnpmi{} for intrinsic evaluations.

	\begin{definition}[Crosslingual Normalized Pointwise Mutual Information, \cnpmi{}]
		Let $\mathcal{W}^{(\ell_1,\ell_2)}_C$ be the set of top $C$ words in a bilingual topic,
		and $\mathcal{R}^{(\ell_1,\ell_2)}$ a parallel reference corpus.
		The \cnpmi{} of this topic is calculated as
		\begin{align}
			\cnpmi{}\left(\mathcal{W}^{(\ell_1,\ell_2)}_C\right) ~=~ -\frac{1}{C^2}\sum_{w_{i},w_{j}\in\mathcal{W}^{(\ell_1,\ell_2)}_C}\frac{\log \frac{\Pr\left(w_{i},w_{j}\right)}{\Pr\left(w_{i}\right)\Pr\left(w_{j}\right)}}{\log\Pr\left(w_{i},w_{j}\right)},
		\end{align}
		where $w_i$ and $w_j$ are from languages $\ell_1$ and $\ell_2$ respectively.
		Let $\mathbf{d}=\left(d_{\ell_1},d_{\ell_2}\right)$ be a pair
		of parallel documents from the reference corpus $\mathcal{R}^{(\ell_1,\ell_2)}$,
		whose size is denoted as
		$\left|\mathcal{R}^{(\ell_1,\ell_2)}\right|$.
		$\left|\left\lbrace{\mathbf{d}}: w_{i}\in{d}_{\ell_1},w_{j}\in{d}_{\ell_2}\right\rbrace\right|$
		is the number of parallel document pairs in which $w_i$ and $w_j$ appear.
		The co-occurrence probability of a word pair and the probability of a single word are calculated as
		\begin{align}
			\Pr\left(w_i,w_j\right) &~\triangleq~   \frac{\left|\left\lbrace{\mathbf{d}}: w_{i}\in{d}_{\ell_1},w_{j}\in{d}_{\ell_2}\right\rbrace\right|}{\left|\mathcal{R}^{(\ell_1,\ell_2)}\right|},\\
			\Pr\left(w_{i}\right) &~\triangleq~  \frac{\left|\left\lbrace{\mathbf{d}}: w_{i}\in{d}_{\ell_1}\right\rbrace\right|}{\left|\mathcal{R}^{(\ell_1,\ell_2)}\right|}.
		\end{align}
	\end{definition}

\begin{figure}
		\centering
		\includegraphics[width=\linewidth]{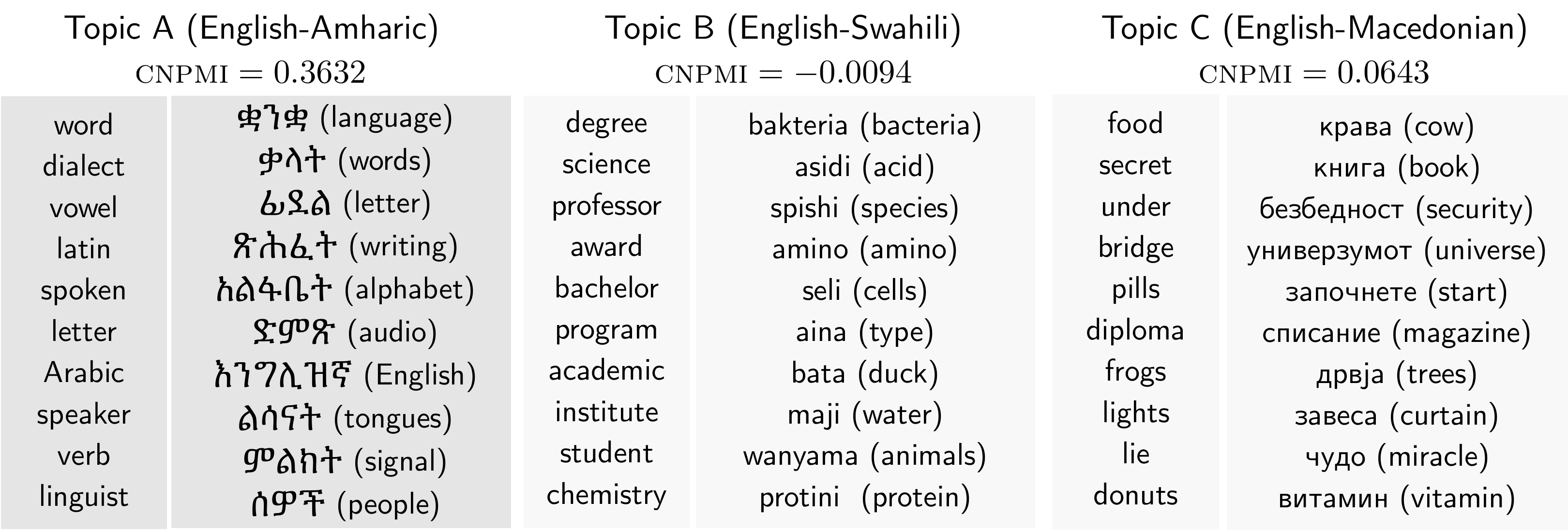}
		\caption{\cnpmi{} measures how likely a bilingual word pair appears in a similar context in two languages, provided by a reference corpus. Topic A has a high \cnpmi{} score since both languages are talking about the same theme. Both Topic B and Topic C are incoherent multilingual topics, although Topic B is coherent within each language.}
		\label{fig:cnpmiexample}
	\end{figure}

	Intuitively,
	a coherent topic should contain words that make sense or fit in a specific context together.
	In the multilingual case,	
	\cnpmi{} measures how likely it is that a bilingual word pair appears in a similar context
	provided by the parallel reference corpus.
	We provide toy examples in Figure~\ref{fig:cnpmiexample},
	where we show three bilingual topics.
	In Topic A, both languages are about ``language'', and all the bilingual word pairs
	have high probability of appearing in the same comparable document pairs.
	Thus Topic A is coherent \textit{crosslingually}, and thus expected to have a high \cnpmi{} score.
	Although we can identify the themes within each language in Topic B,
	\textit{i.e.,} education in English and biology in Swahili,
	most of the bilingual word pairs do not make sense or appear in the same context,
	which gives us a low \cnpmi{} score.
	The last topic is not coherent even within each language,
	so it has low \cnpmi{} as well.
	Through this example,
	we see that \cnpmi{} detects crosslingual coherence in multiple ways,
	unlike other intrinsic measures that might be adapted for crosslingual models.

	In our experiments,
	we use $10{,}000$ linked Wikipedia article pairs for each language pair $(\en{},\ell)$ ($20{,}000$ in total) as the reference corpus, and set $C=10$ by default.
	Note that \highlan{} has more Wikipedia articles,
	and we make sure the articles used for evaluating \cnpmi{} scores
	do not appear in the training set.
	However, for \lowlan{},
	since the number of linked Wikipedia articles is extremely limited,
	we use all the available pairs to evaluate \cnpmi{} scores.
	The statistics are shown in Table~\ref{evalset}.
	
	\begin{table}\centering
		\begin{tabular}{c|c|r|r|r|r|r|r} \hline
			& & \multicolumn{3}{c|}{ English } & \multicolumn{3}{c}{ Paired language } \\ \hline
			& & \#docs & \#tokens & \#types & \#docs & \#tokens & \#types \\ \hline \hline
			\multirow{5}{*}{\highlan}
			&\ar{} & 10,000 & 3,597,322 & 128,926 & 10,000 & 996,801 & 64,197 \\ \cline{2-8}
			&\de{} & 10,000 & 2,155,680 & 103,812 & 10,000 & 1,459,015 & 166,763 \\ \cline{2-8}
			&\es{} & 10,000 & 3,021,732 & 149,423 & 10,000 & 1,737,312 & 142,086 \\ \cline{2-8}
			&\ru{} & 10,000 & 3,016,795& 154,442& 10,000 & 2,299,332 & 284,447 \\ \cline{2-8}
			&\zh{} & 10,000 & 1,982,452 &  112,174& 10,000 & 1,335,922 & 144,936 \\ \hline
			\multirow{5}{*}{\lowlan}
			& \am{} & 4,316 & 9,632,700 & 269,772 & 4,316 & 403,158 & 91,295\\ \cline{2-8}
			& \ay{} & 4,187 & 5,231,260 & 167,531 & 4,187 & 280,194 & 32,424\\ \cline{2-8}
			& \mk{} & 10,000 & 11,080,304 & 301,026 & 10,000 & 3,175,182 & 245,687\\ \cline{2-8}
			& \sw{} & 10,000 & 13,931,839 & 341,231 & 10,000 & 1,755,514 & 134,152\\ \cline{2-8}
			& \tl{} & 6,471 & 7,720,517 & 645,534 & 6,471 & 1,124,049 & 83,967 \\ \hline
		\end{tabular} 
		\caption{Statistics of Wikipedia corpus for topic coherence evaluation (\cnpmi{}).}
		\label{evalset}
	\end{table}

	\subsubsection{Extrinsic Evaluation: Crosslingual Classification}
	\label{sec:ex}
	
	Crosslingual document classification is the most common downstream application for multilingual topic models~\cite{HeymanVM16,VulicSTM15,SmetTM11}.
	Typically, a model is trained on a multilingual training set $D^{(\ell_1,\ell_2)}$
	in languages $\ell_1$ and $\ell_2$.
	Using the trained topic-vocabulary distributions $\phi$,
	the model infers topics in test sets ${D'}^{(\ell_1)}$ and ${D'}^{(\ell_2)}$.
	
\begin{figure}
		\centering
		\includegraphics[width=\linewidth]{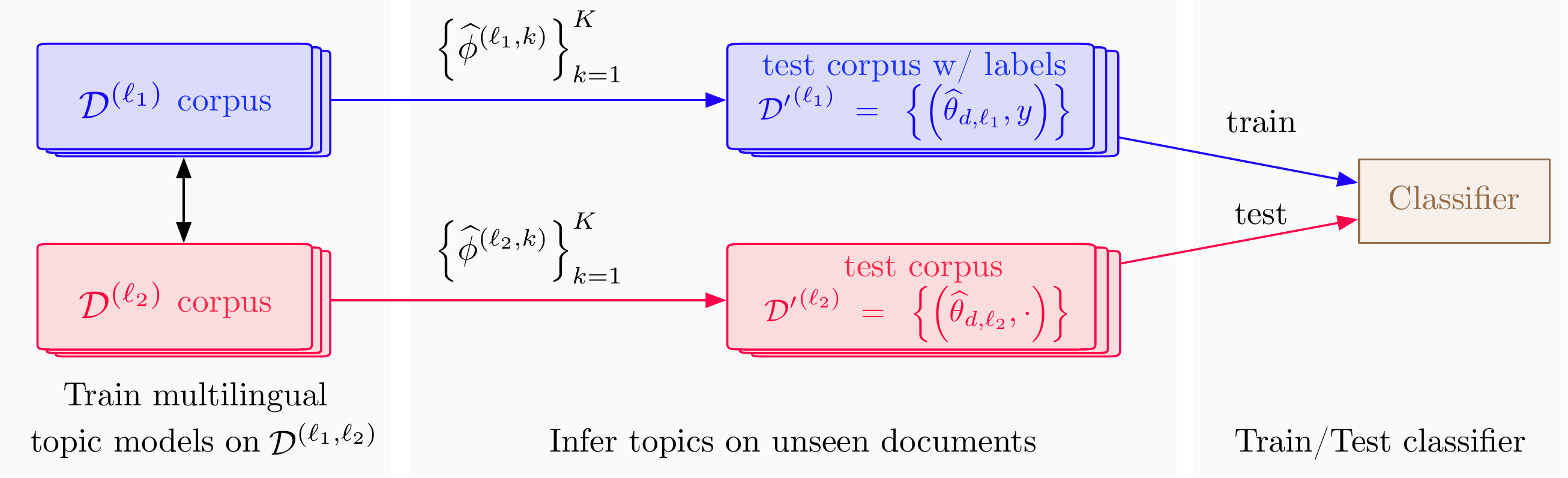}
		\caption{An illustration of crosslingual document classification. After training multilingual topic models,
			the topics, $\{\widehat{\phi}^{(\ell,k)}\}$ are used to infer document-topic distributions $\widehat{\theta}$
			of unseen documents in both languages. A classifier is trained with the inferred distributions $\widehat{\theta}_{d,\ell_1}$ as features and the labels $y$ in language $\ell_1$,
			and predicts labels in language $\ell_2$.
		}
		\label{fig:mlc}
	\end{figure}

	In multilingual topic models,
	document-topic distributions $\theta$ can be used as features for classification,
	where the $\widehat{\theta}_{d,\ell_1}$ vectors in language $\ell_1$ train a classifier
	tested by the $\widehat{\theta}_{d,\ell_2}$ vectors in language $\ell_2$.
	A better classification performance indicates
	more consistent features across languages.
	See \fig{fig:mlc} for an illustration.
	In our experiments,
	we use a linear support vector machine (SVM) to train multilabel classifiers with five-fold cross-validation.
	Then, we use micro-averaged F-1 scores to evaluate and compare performance across different models.
	
	For crosslingual classification, we also require held-out test data with labels or annotations.
	In our experiments,
	we construct test sets from two sources: 
	TED Talks 2013 (\ted{}) and Global Voices (\gv{}).
	\ted{} contains parallel documents in all languages in \highlan{},
	while \gv{} contains all languages from both \highlan{}
	and \lowlan{}.
	
	Using the two multilingual sources,
	we create two types of test sets for \highlan{}---\ted{}+\ted{} and \ted{}+\gv{},
	and only one type for \lowlan{}---\ted{}+\gv{}.
	In \textbf{\ted{}+\ted{}},
	we infer document-topic distributions on documents from \ted{} in English and the paired language.
	This only applies to \highlan{},
	since \ted{} do not have documents in \lowlan{}.
	In \textbf{\ted{}+\gv{}},
	we infer topics on English documents from \ted{},
	and infer topics on documents from \gv{} in the paired language (both \highlan{} and \lowlan{}).
	The two types of test sets also represent different application situations.
	\ted{}+\ted{} implies that the test documents in both languages are parallel
	and come from the same source,
	while \ted{}+\gv{} represents how the topic model performs
	when the two languages have different data sources.
	
	Both corpora are retrieved from \url{http://opus.nlpl.eu/}~\cite{Tiedemann12}.
	The labels, however, are manually retrieved from \url{http://ted.com/}
	and \url{http://globalvoices.org/}.
	In \ted{} corpus, each document is a transcript of a talk
	and is assigned to multiple categories on the webpage,
	such as ``technology'', ``arts'', and so forth.
	We collect all categories for the entire \ted{} corpus,
	and use the three most frequent categories---\textit{technology}, \textit{culture}, \textit{science}---as document labels.
	Similarly, in \gv{} corpus,
	each document is a news story,
	and has been labeled with multiple categories on the webpage of the story.
	Since in \ted{}+\gv{}, the two sets are from different sources,
	and training and testing is only possible when both sets share the same labels,
	we apply the same three labels from \ted{} to \gv{} as well.
	This processing requires minor mappings,
	\textit{e.g.,} from ``arts-culture'' in \gv{} to ``culture'' in \ted{}.
	The data statistics are presented in Table~\ref{gv}.

	\begin{table}[]\centering
		\begin{tabular}{c|c|r|r|r|r|r|r}
			\hline
			& &\multicolumn{3}{c}{Corpus statistics} & \multicolumn{3}{|c}{Label distributions}\\ \hline
			& & \#docs & \#types & \#tokens & \#\textit{technology} & \textit{culture} & \textit{science} \\ \hline\hline
			\multirow{5}{*}{\ted{}}
			&\ar{} & 1,112 & 1,066,754 & 15,124  & 384 & 304 & 290 \\ \cline{2-8}
			&\de{} & 1,063 & 774,734 & 19,826  & 364 & 289 & 276 \\ \cline{2-8}
			&\es{} & 1,152 & 933,376 & 13,088  & 401 & 312 & 295 \\  \cline{2-8}
			&\ru{} & 1,010 & 831,873 & 17,020  & 346 & 275 & 261 \\ \cline{2-8}
			&\zh{} & 1,123 & 1,032,708 & 19,594  & 386 & 315 & 290 \\ \hline
			\multirow{5}{*}{\shortstack{\gv{}\\ (\highlan{})}}
			&\ar{} & 2,000 & 325,879 & 13,072 & 510 & 489 & 33 \\ \cline{2-8}
			&\de{} & 1,481 & 269,470 & 16,031 & 346 & 344 & 42 \\ \cline{2-8}
			&\es{} & 2,000 & 367,631 & 11,104 & 457 & 387 & 38 \\  \cline{2-8}
			&\ru{} & 2,000 & 488,878 & 16,157 & 516 & 369 & 62 \\ \cline{2-8}
			&\zh{} & 2,000 & 528,370 & 18,194 & 499 & 366 & 56 \\ \hline
			\multirow{5}{*}{\shortstack{\gv{}\\ (\lowlan{})}}
			&\am{} & 39 & 10,589 & 4,047 & 3 & 3 & 1 \\ \cline{2-8}
			&\ay{} & 674 & 66,076 & 4,939 & 76 & 100 & 46 \\ \cline{2-8}
			&\mk{} & 1,992 & 388,713 & 29,022 & 343 & 426 & 182 \\  \cline{2-8}
			&\sw{} & 1,383 & 359,066 & 14,072 & 137 & 110 & 71 \\ \cline{2-8}
			&\tl{} & 254 & 26,072 & 6,138 & 32 & 67 & 19 \\ \hline
		\end{tabular}
		\caption{Statistics of TED Talks 2013 (\ted{}) and Global Voices (\gv{}) corpus.}
		\label{gv}
	\end{table}

	\section{Document-Level Transfer and Its Limitations}
	\label{ex:ex}
	
	We first explore the empirical characteristics of document-level transfer,
	using \doclink{}, \cbilda{} and \softlink{}.
	
	Multilingual corpora can be loosely categorized into three types---parallel,
	comparable, and incomparable.
	A parallel corpus contains exact document translations across languages,
	of which \textsc{EuroParl} and The Bible we discussed before are examples.
	A comparable corpus contains document pairs (in the bilingual case),
	where each document in one language has a related counterpart in the other language.
	However, these document pairs are not exact translations of each other,
	and they can only be connected through a loosely defined ``theme''.
	Wikipedia is an example,
	where document pairs are linked by article titles.
	Incomparable corpora contain potentially unrelated documents across  languages, with no explicit indicators of document pairs.
	
	With different levels of comparability 
	comes different availabilities of such corpora:
	it is much harder to find parallel corpora in low-resource languages.
	Therefore,
	we first focus on \highlan{},
	and use Wikipedia to simulate the low-resource situation in Section~\ref{sec:senscorpus},
	where we find that \doclink{} and \cbilda{} are very sensitive to the training corpus,
	and thus might not be the best option when it comes to low-resource languages.
	We then examine \lowlan{} in Section~\ref{sec:lowlan}

	\subsection{Sensitivity to Training Corpus} \label{sec:senscorpus}
	
	\begin{figure}
		\includegraphics[width=\linewidth]{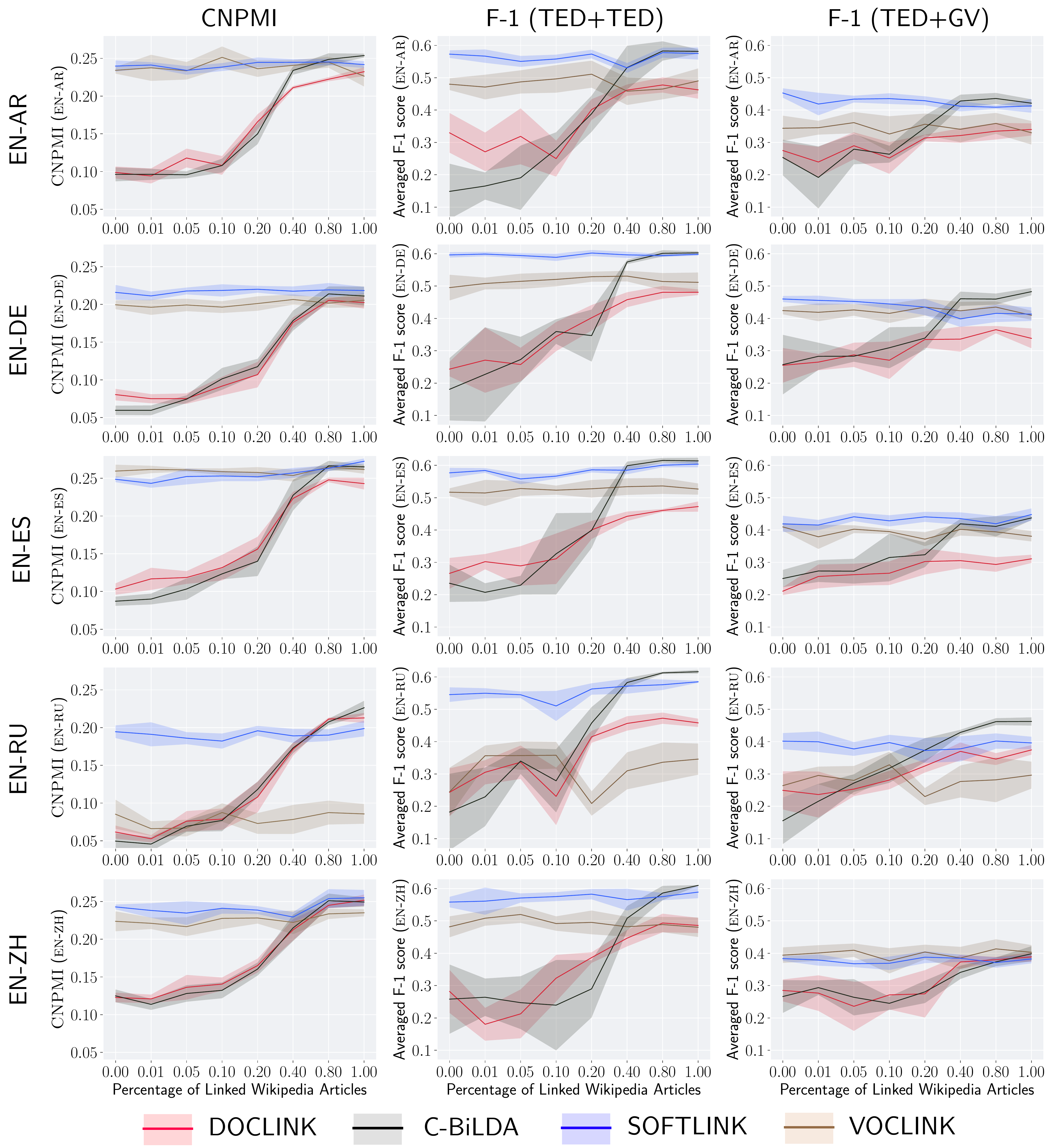}
		\caption{Both \softlink{} and \voclink{} stay at a stable performance level of either \cnpmi{} or F-1 scores, while \doclink{} and \cbilda{} expectedly have better performance as there are more linked Wikipedia articles.}
		\label{fig:trend}
	\end{figure}
	
	We first vary the comparability of the training corpus
	and study how different models behave under different situations.
	All models are potentially affected by the comparability of the training set,
	although only \doclink{} and \cbilda{} explicitly rely on this information to define transfer operation.
	This experiment shows that
	models transferring knowledge on document level (\doclink{} and \cbilda{}) are very sensitive to the training set,
	but can be almost entirely insensitive with appropriate modifications to the transfer operation, \textit{i.e.,} with \softlink{}.

	\subsubsection{Experiment Settings}
	For each language pair $(\en{}, \ell)$,
	we construct a random subsample of $2{,}000$ documents from Wikipedia in each language ($4{,}000$ in total).
	To vary the comparability,
	we vary the proportion of linked Wikipedia articles between the two languages,
	from $0.0$, $0.01$, $0.05$, $0.1$, $0.2$, $0.4$, $0.8$, to $1$.
	When the percentage is zero,
	the bilingual corpus is entirely \textit{incomparable}---no document-level translations can be found in another language,
	and \doclink{} and \cbilda{} degrade into monolingual LDAs.
	The indicator matrix used by transfer operation in Section~\ref{ex:doclink}
	is a zero matrix $\delta=\mathbf{0}$.
	When the percentage is one,
	meaning each document from one language is linked to one document from another language,
	the corpus is considered fully comparable,
	and $\delta$ is an identity matrix $\mathbf{1}$.
	Any number between $0$ and $1$ makes the corpus \textit{partially comparable} to different degrees.
	The \cnpmi{} and crosslingual classification results are shown in Figure~\ref{fig:trend},
	and the shades indicate the standard deviations across five Gibbs sampling chains.
	For \voclink{} and \softlink{},
	we use all the dictionary entries.

	\subsubsection{Results}
	In terms of topic coherence (\cnpmi{}),
	both \doclink{} and \cbilda{} have competitive performance on \cnpmi{},
	and achieve full potential when the corpus is fully comparable.
	As expected, 
	models transferring knowledge at the document level (\doclink{} and \cbilda{})
	are very sensitive to 
	the training corpus:
	the more aligned the corpus is, the better topics the model learns.
	For the word-level model,
	\voclink{} roughly stays at the same performance level
	which is also expected, since this model does not use linked documents as supervision.
	However, its performance on Russian is surprisingly low compared to other languages and models.
	In the next section,
	we will look closer at this problem by investigating the impact of dictionaries.
	
	It is notable that \softlink{}, a document-level model,
	is also insensitive to the training corpus
	and outperforms other models most of the time.
	Recall that on document level,
	\softlink{} defines transfer operation on document-topic distributions $\theta$,
	similar as \doclink{} and \cbilda{},
	but using dictionary resources.
	This implies that good design of the supervision $\delta$ in the transfer operation
	could lead to a more stable performance across different training situations.
	
	When it comes to the classification task,
	the F-1 scores of \doclink{} and \cbilda{} have very large variations,
	and the increasing trend of F-1 scores is less obvious than with \cnpmi{}.
	This is especially true when the percentage of linked documents is very small.
	For one, when the percentage is small,
	the transfer on the document level is less constrained,
	leaving the projection of two languages into the same topic space less predictive.
	The evaluation scope of \cnpmi{} is actually much smaller and more concentrated than classification,
	since it only focuses on the top $C$ words,
	which does not lead to large variations.

	One consistent result we notice is that
	\softlink{} still performs well on classification
	with very small variations and stable F-1 scores,
	which again benefits from the definition of transfer operation in \softlink{}.
	When transferring topics to another language,
	\softlink{} uses dictionary constraints as in \voclink{},
	but instead of a simple one-on-one word type mapping,
	it expands the transfer scope to the entire document.
	Additionally,
	\softlink{} distributionally transfers knowledge from the entire corpus in another language,
	which actually reinforces the transfer efficiency
	without relying on direct supervision at the document level.

	\begin{figure}[!ht]
		\centering
		\begin{subfigure}[t]{\linewidth}
			\centering
			\includegraphics[width=\linewidth]{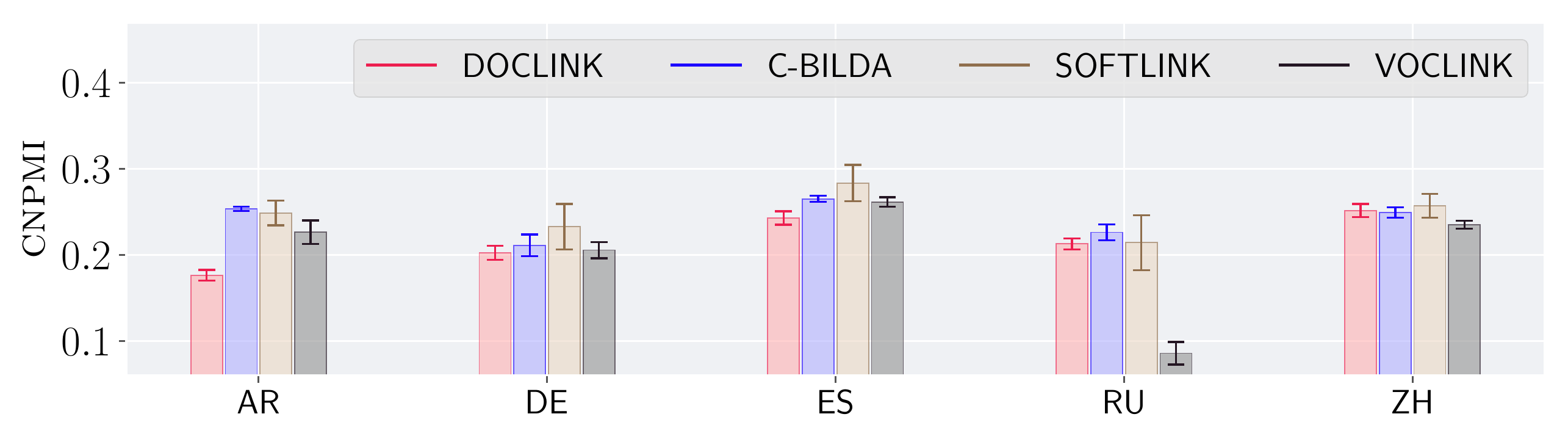}
			\includegraphics[width=\linewidth]{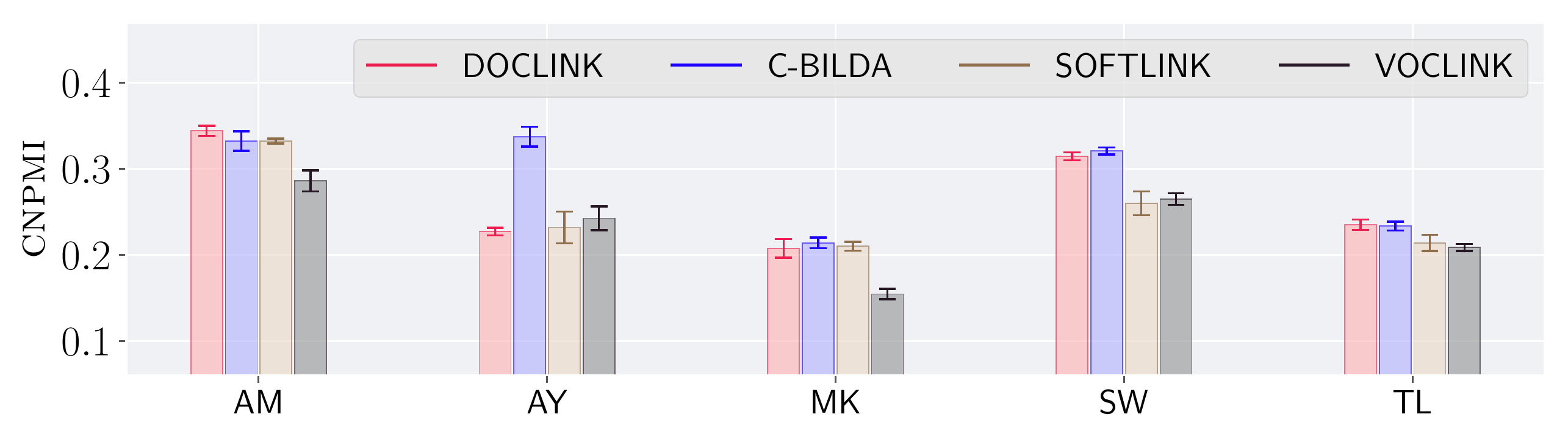}
			\caption{\cnpmi{} score comparison of different models and languages with cardinality $C=10$.}
			\label{lowlancnpmi}
		\end{subfigure}
		
		\begin{subfigure}[t]{\linewidth}
			\centering
			\includegraphics[width=\linewidth]{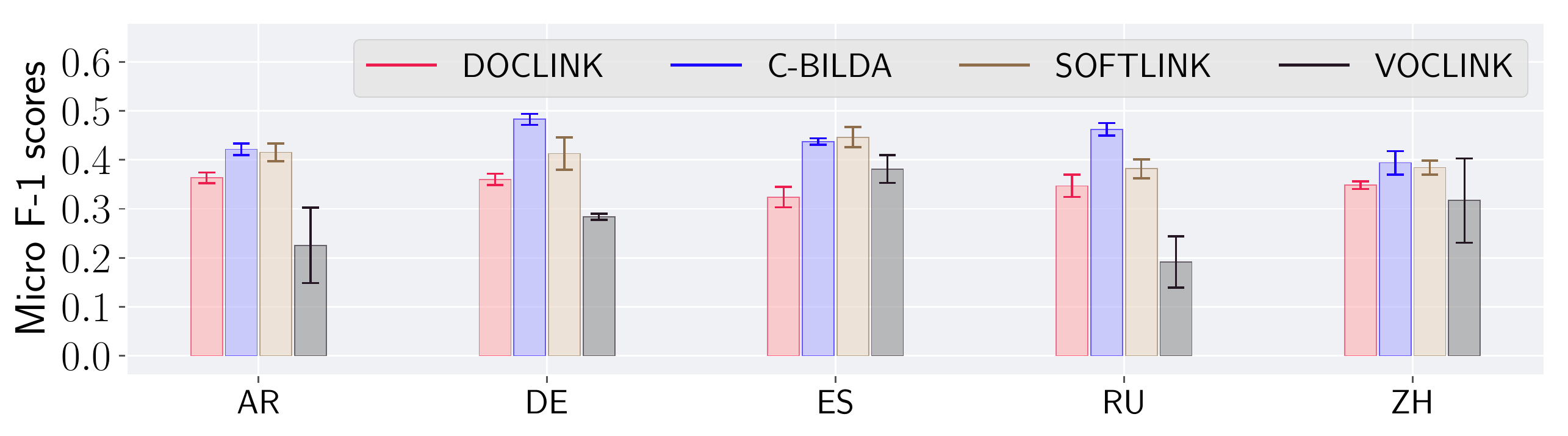}
			\includegraphics[width=\linewidth]{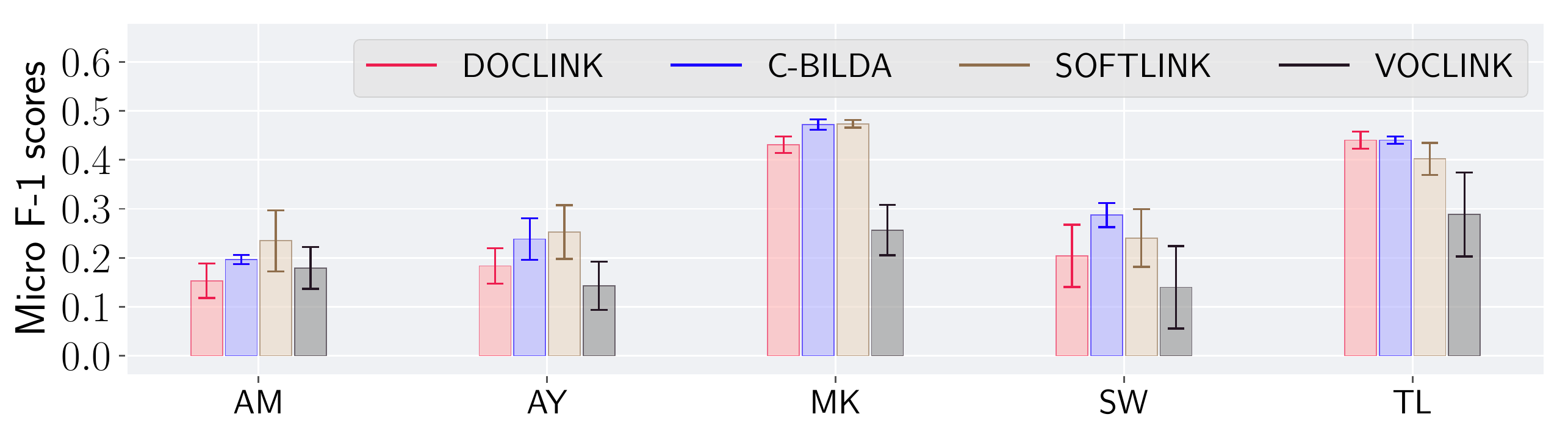}
			\caption{Micro-averaged F-1 scores of different models and languages on \ted{}+\gv{} corpora. }
			\label{my-label}
		\end{subfigure}
		
		\caption{Topic quality evaluation and classification performance on both \highlan{} and \lowlan{}. We notice that \voclink{} has lower \cnpmi{} and F-1 scores in general, with large standard deviations. \cbilda{}, on the other hand, outperforms other models in most of the languages.}
	\end{figure}

	\subsection{Performance on \lowlan{}}\label{sec:lowlan}

	In this section,
	we take a look at languages in \lowlan{}.
	For \softlink{} and \voclink{}, we use all
	dictionary entries to train languages in \lowlan{},
	since the sizes of dictionaries in these languages are already very small.
	We again use a subsample of $2{,}000$ Wikipedia document pairs with English
	to make the results comparable with \highlan{}.
	In Figure~\ref{lowlancnpmi},
	we also present results of models for \highlan{} using fully comparable training corpora and full dictionaries for direct comparison of the effect of language resources.

	In most cases, transfer on document level (particularly \cbilda{}) performs better than on word levels,
	in both \highlan{} and \lowlan{}.
	Considering the number of dictionary entries available from Table~\ref{wiki},
	it is reasonable to suspect that the dictionary is a major factor
	affecting the performance of word-level transfer.
	
	On the other hand,
	although \softlink{} does not model vocabularies directly as in \voclink{},
	transferring knowledge at the document level with a limited dictionary
	still yields competetive \cnpmi{} scores.
	Therefore, in this experiment on \lowlan{},
	we see that with the same lexicon resource,
	it is generally more efficient to transfer knowledge on document level.
	We will also explore this in detail in Section~\ref{wt}.

	We also present a comparison of micro-averaged F-1 scores between \highlan{} and \lowlan{} in \fig{my-label}.
	The test set used for this comparison is $\ted{}+\gv{}$,
	since \ted{} does not have articles available in \lowlan{}.
	Also, languages such as Amharic (\am{}) have fewer than $50$ \gv{} articles available,
	which is an extremely small number for training a robust classifier,
	so in these experiments,
	we only train classifiers on English (\ted{} articles) and test them on languages in \highlan{} and \lowlan{} (\gv{} articles).
	
	Similarly, the classification results are generally better in document-level transfer,
	and both \cbilda{} and \softlink{} give similar scores.
	However, it is worth noting that \voclink{} has very large
	variations in all languages,
	and the F-1 scores are very low.
	This suggests again that transferring knowledge on the word level is less effective,
	and in Section~\ref{wt} we study in detail why this is the case.

	\section{Word-Level Transfer and Its Limitations}
	\label{wt}
	
	In the previous section, we compared different multilingual topic models
	with a focus on document-level models.
	We draw conclusions that \doclink{} and \cbilda{}
	are very sensitive to the training corpus, which is naturally due to their definition of
	supervision as a one-to-one document pair mapping.
	On the other hand, the word-level model \voclink{}
	in general has lower performance, especially with \lowlan{},
	even if the corpus is entirely comparable.
	
	One interesting result we observed from the previous section
	is that \softlink{} and \voclink{} use the same dictionary resource
	while transferring topics on different levels,
	and \softlink{} generally has better performance than \voclink{}.
	Therefore,
	in this section,
	we explore the characteristics of the word-level model \voclink{} and compare it with \softlink{}
	to study why it doesn't use the same dictionary resource as effectively.
	
	To this end,
	we first vary the amount of dictionary entries available
	and compare how \softlink{} and \voclink{} perform (Section~\ref{sec:sensed}).
	Based on the results,
	we analyze word-level transfer from three different angles:
	dictionary usage (Section~\ref{sec:dictvoc}) as an intuitive explanation of the models,
	topic analysis (Section~\ref{sec:topicanalysis}) from a more qualitative perspective,
	and comparing transfer strength (Section~\ref{sec:comparestrength}) as a quantitative analysis.
	
	\begin{figure}
		\centering
		\includegraphics[width=\linewidth]{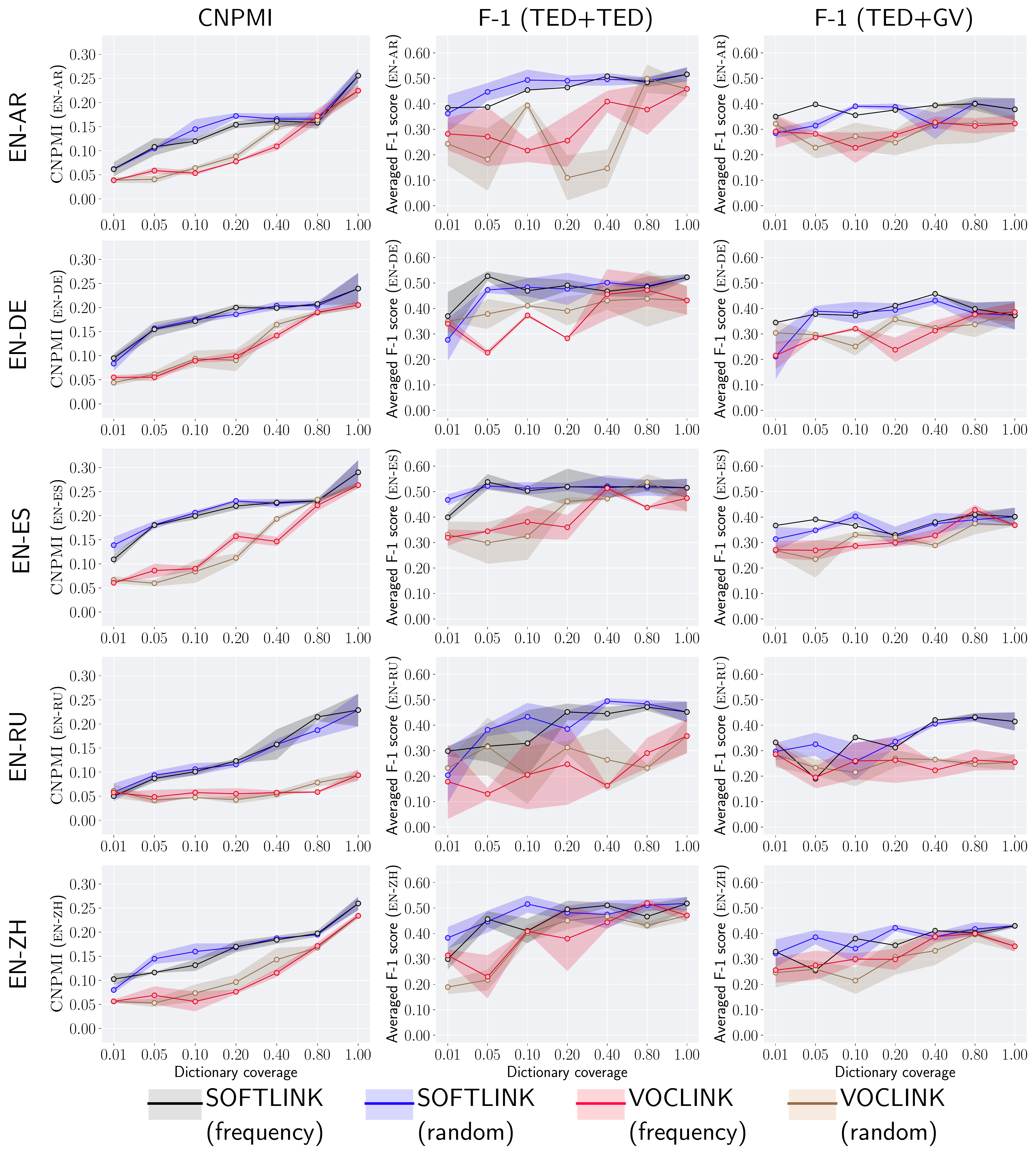}
		\caption{\softlink{} produces better topics and is more capable of crosslingual classification tasks than \voclink{} when the number of dictionary entries is very limited.}
		\label{fig:dictionary}
	\end{figure}

	\subsection{Sensitivity to Dictionaries}
	\label{sec:sensed}
	
	Word-level models such as \voclink{} use a dictionary as supervision,
	and thus will naturally be affected by the dictionary used.
	Although \softlink{} transfers knowledge on the document level,
	it uses dictionary to calculate the transfer distributions used in its document-level transfer operation.
	In this section,
	we focus on the comparison of \softlink{} and \voclink{}.

	\subsubsection{Sampling the Dictionary Resource}
	The dictionary is the essential part of \softlink{} and \voclink{}
	and is used in different ways to define transfer operations.
	The availability of dictionaries, however,
	varies among different languages.
	From \tbl{wiki},
	we notice that for \lowlan{} the number of available dictionary entries
	is very limited,
	which suggests it could be a major factor affecting the performance of word-level topic models.
	Therefore, in this experiment,
	we sample different numbers of dictionary entries in \highlan{}
	to study how this alters performance of \softlink{} and \voclink{}.
	
	Given a bilingual dictionary,
	we add only a proportion of entries in it to \softlink{} and \voclink{}.
	As in the previous experiments varying the proportion of document links,
	we change the proportion from $0$, $0.01$, $0.05$, $0.1$, $0.2$, $0.4$, $0.8$, to $1.0$.
	When the proportion is $0$,
	both \softlink{} and \voclink{} become monolingual LDA and no transfer happens;
	when the proportion is $1$,
	both models reach their highest potential with all the dictionary entries available.
	
	We also sample the dictionary in two manners: random- and frequency-based.
	In random-based,
	the entries are randomly chosen from the dictionary,
	and the five chains are having different entries added to the models.
	In frequency-based,
	we select the most frequent word types from the training corpus.

	Figure~\ref{fig:dictionary} shows a detailed comparison among different evaluations and languages.
	As expected,
	adding more dictionary entries helps both \softlink{} and \voclink{}, with
	increasing \cnpmi{} scores and F-1 scores in general.
	However, we notice that adding more dictionary entries can boost \softlink{}'s performance very quickly,
	while the increase in \voclink{}'s \cnpmi{} scores is slower.
	Similar trends can be observed in the classification task as well,
	where adding more words does not necessarily increase \voclink{}'s F-1 scores,
	and the variations are very high.
	
	This comparison provides an interesting insight to increasing lexical resources efficiently.
	In some applications, especially related to low-resource languages,
	the number of available lexicon resources is very small,
	and one way to solve this problem is to incorporate human feedback,
	such as interactive topic modeling proposed by~\namecite{HuBSS14}.
	In our case,
	a native speaker of the low-resource language could provide word translations
	that could be incorporated into topic models.
	Due to limited time and financial budget, however,
	it is impossible to translate all the word types that appear in the corpus,
	so the challenge is how to boost the performance of the target task as much as possible
	with less effort from humans.
	In this comparison,
	we see that if the target task is to train coherent multilingual topics, 
	training \softlink{} is a more efficient way than \voclink{}.

	\subsubsection{Varying Comparability of the Corpus}

	For \softlink{} and \voclink{},
	the dictionary is only one aspect of the training situation.
	As discussed in our document-level experiments,
	the training corpus is also an important factor that could affect
	the performance of all topic models.
	While corpus comparability is not an explicit requirement of \softlink{} and \voclink{}, the comparability of the corpus might affect the coverage provided by the dictionary or affect performance in other ways.
	In \softlink{}, comparability could also affect the transfer operator's ability to find similar documents to link to.
	In this section,
	we study the relationship between dictionary coverage and comparability of the training corpus.

	\begin{figure*}
		\centering
		\begin{subfigure}[t]{\textwidth}
			\includegraphics[width=\linewidth]{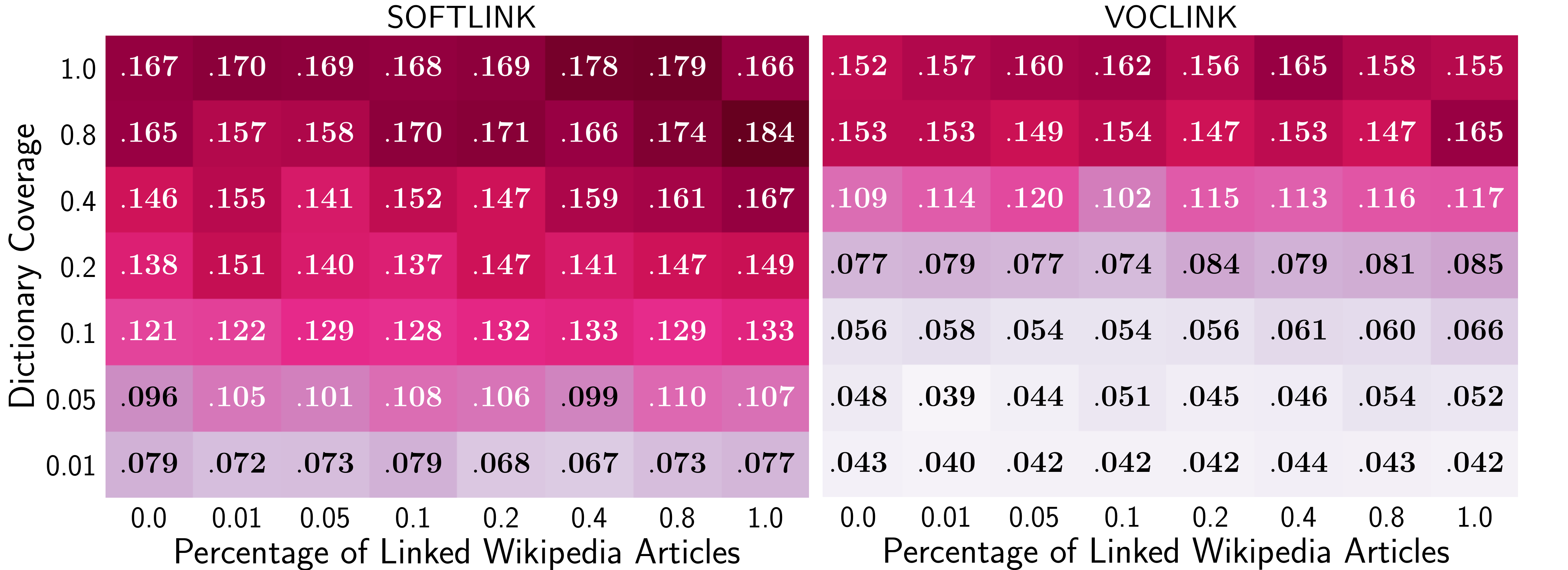}
			\caption{Average \cnpmi{} scores on multilingual topic coherence.}
			\label{fig:mlchigh-dict}
		\end{subfigure}
		\begin{subfigure}[t]{\textwidth}
			\includegraphics[width=\linewidth]{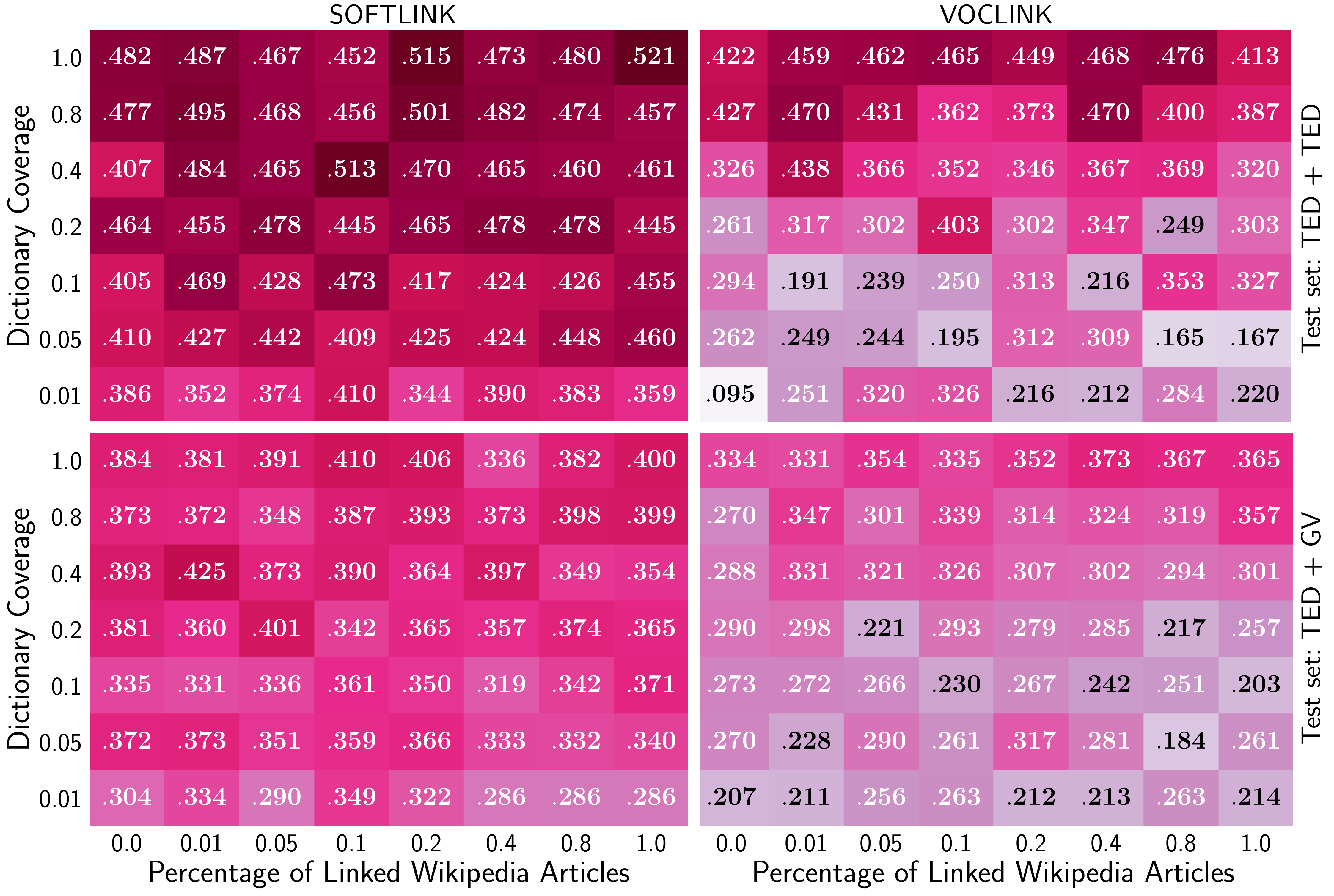}
			\caption{Multilabel crosslingual document classification F-1 scores in \highlan{}.}
			\label{fig:mlchigh}
		\end{subfigure}
		\caption{Adding more dictionary entries has higher impact on word-level model \voclink{}. \softlink{} learns better quality topics than \voclink{}. \softlink{} also generally performs better on classification.}
	\end{figure*}

	Similar to the previous section,
	we vary the dictionary coverage
	from $0.01$, $0.05$, $0.1$, $0.2$, $0.4$, $0.8$, to $1$,
	using the frequency-based method as in the last experiment.
	We also vary the number of linked Wikipedia articles
	from $0$, $0.01$, $0.05$, $0.1$, $0.2$, $0.4$, $0.8$, to $1$.
	We present \cnpmi{} scores in Figure~\ref{fig:mlchigh-dict},
	where the results are averaged over all five languages in \highlan{}.
	It is clear that \softlink{} outperforms \voclink{},
	regardless of training corpus and dictionary size.
	This implies that \softlink{} could potentially learn coherent multilingual topics
	even when the training conditions are unfavorable, \textit{i.e.,}
	the training corpus is incomparable and there is only a small number of dictionary entries.
	
	The results of crosslingual classification are shown in \fig{fig:mlchigh}.
	When the test sets are from the same source (\ted{}+\ted{}),
	\softlink{} utilizes dictionary more efficiently
	and performs better than \voclink{}.
	In particular,
	F-1 scores of \softlink{} using only $20\%$ of dictionary entries
	is already outperforming \voclink{} using the full dictionary.
	Similar comparison can also be drawn when the test sets are from different sources, \textit{i.e.,} \ted{}+\gv{}.

	\subsubsection{Discussion}
	From the results so far,
	it is empirically clear that transferring knowledge on the word level
	tends to be less efficient than the document level.
	This is arguably counter-intuitive.
	Recall that the goal of multilingual topic models
	is to let semantically related words and translations have similar distributions over topics.
	The word-level model \voclink{} directly uses this information, \textit{i.e.,} dictionary entries,
	to define transfer operation,
	yet its \cnpmi{} scores are lower.
	In the following sections, therefore, we try to explain this apparent contradiction.
	We first analyze the dictionary usage of \voclink{} (Section~\ref{sec:dictvoc}),
	and then lead our discussion on the transfer strength comparisons between
	document and word levels for all models (Sections~\ref{sec:topicanalysis} and \ref{sec:comparestrength}).

	\subsection{Dictionary Usage}\label{sec:dictvoc}
	\begin{figure}
		\centering
		\includegraphics[width=\linewidth]{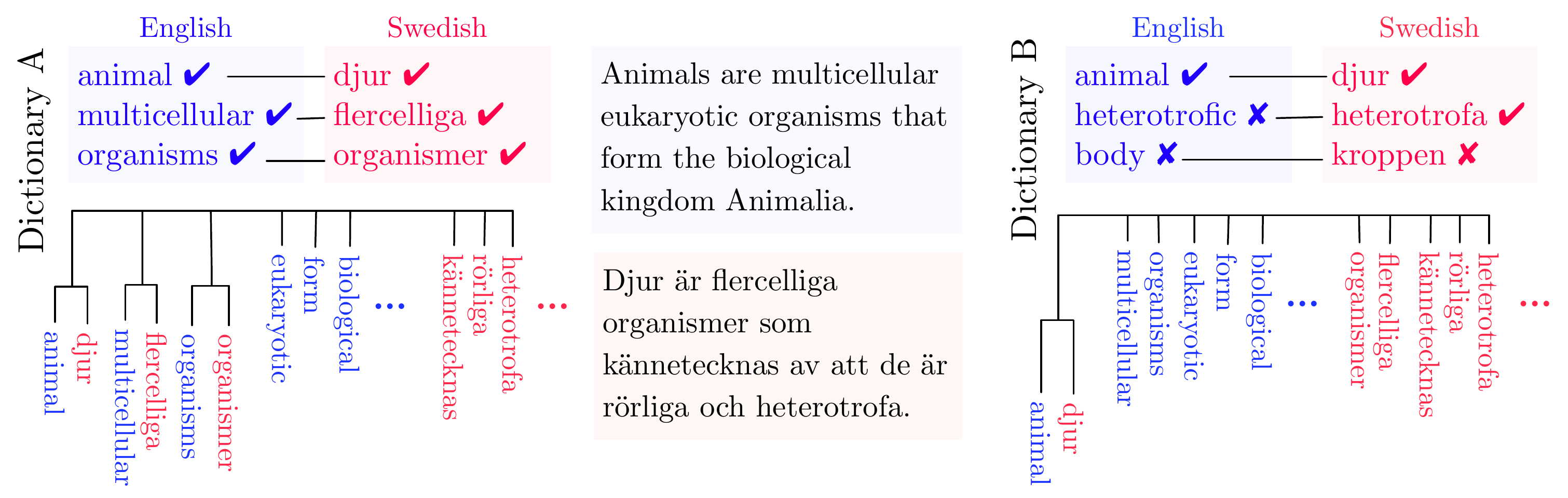}
		\caption{The dictionary used by \voclink{} is affected by its overlap with the corpus.
			In this example, the three entries in Dictionary A can all be found in the corpus,
			so the tree structure has all of them.
			However, only one entry in Dictionary B can be found in the corpus.
			Although the Swedish word ``heterotrofa'' is also in the dictionary,
			its English translation cannot be found in the corpus,
			so Dictionary B ends up a tree with only one entry.
		}
		\label{fig:dictionaryuse}
	\end{figure}

	\begin{figure}
		\centering
		\includegraphics[width=\linewidth]{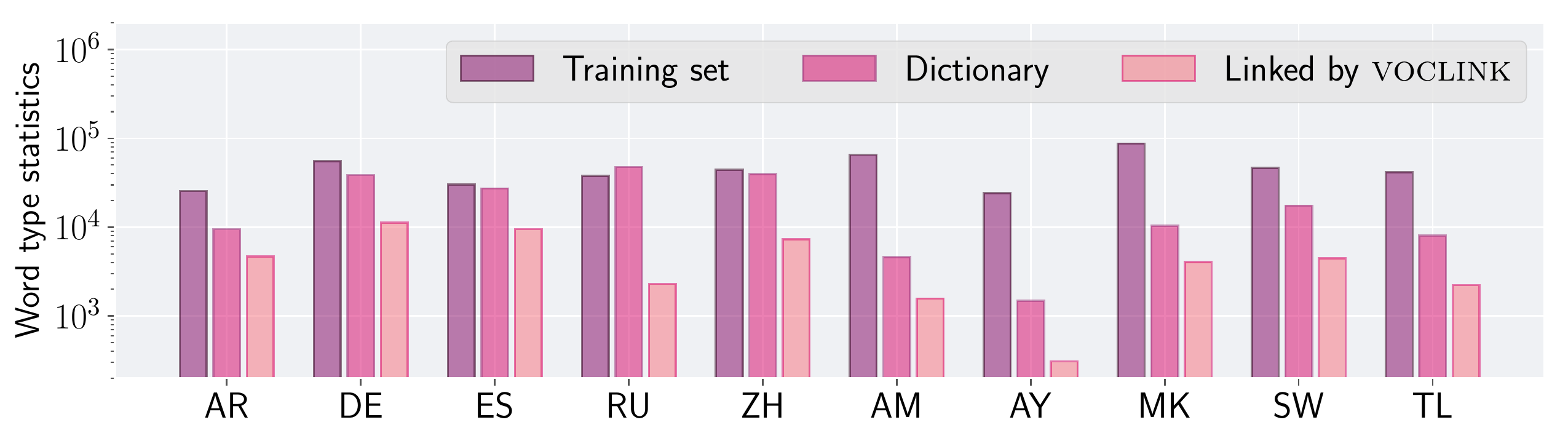}
		\caption{The number of word types that are linked in \voclink{} is far less than the original dictionary and even than that of word types in the training sets.}
		\label{fig:dictcoverage}
	\end{figure}
	In practice,
	the assumption of \voclink{} is also often weakened by another important factor---the presence of word translations in the training corpus.
	Given a word pair $\left(w^{(\ell_1)},w^{(\ell_2)}\right)$,
	the assumption of \voclink{} is valid only when both words appear in the training corpus in their respective languages.
	If $w^{(\ell_2)}$ is not in $D^{(\ell_2)}$,
	$w^{(\ell_1)}$ will be treated as an untranslated word instead.
	Figure~\ref{fig:dictionaryuse} shows an example of how tree structures in \voclink{}
	are affected by the corpus and the dictionary.

	In \fig{fig:dictcoverage},
	we present the statistics of word types from different sources on a logarithmic scale.
	``Dictionary'' is the number of word types that appeared in the original dictionary
	as shown in the last column of Table~\ref{wiki},
	and we use the same preprocessing to the dictionary as to the training corpus to make sure the quantities
	are comparable.
	``Training set'' is the number of word types that appeared in the training set,
	and ``Linked by \voclink{}'' is the number of word types that are actually used
	in \voclink{}, \textit{i.e.,} the number of non-zero entries in $\delta$ in the transfer operation.
	
	Note that even when we use the complete dictionary to create the tree structure in \voclink{},
	in \lowlan{},  there are far more word types in the training set than those in the dictionary.
	In other words,
	the supervision matrix $\delta$ used by $h_{\phi^{(r,k)}}$ is never actually full rank,
	and thus,
	the full potential of \voclink{} is very difficult to achieve
	due to the properties of the training corpus.
	This situation is as if the document-level model \doclink{} had only half of the linked documents in the training corpus.
	
	On the other hand,
	we notice that in \highlan{},
	the number of word types in the dictionary is usually comparable to that of the training set (except in \ar{}).
	However, for \lowlan{},
	the situation is quite the contrary,
	\textit{i.e.,} there are more word types in the training set than in the dictionary.
	Thus, the availability of sufficient dictionary entries is especially a problem for \lowlan{}.

	We conclude from \fig{fig:mlchigh-dict} that
	adding more dictionary entries will slowly improve \voclink{},
	but even when there are enough dictionary items,
	due to model assumptions, \voclink{} will not achieve its full potential unless every word in the training corpus is in the dictionary.
	A possible solution is to first extract word alignments from parallel corpora,
	and then create a tree structure using those word alignments, as experimented in~\namecite{HuZEB14}.
	However, when parallel corpora are available, we have shown that document-level models such as \doclink{} work better anyway,
	and the accuracy of word aligners is another possible limitation to consider.

	\subsection{Topic Analysis}\label{sec:topicanalysis}
	
	While \voclink{} uses a dictionary to directly model word translations,
	\softlink{} uses the same dictionary to define the supervision in transfer operation differently
	on the document level.
	Experiments show that transferring knowledge on the document level with a dictionary,
	\textit{i.e.,} \softlink{}, is more efficient, resulting in
	stable and low-variance topic qualities in various training situations.
	A natural question is why the same resource results in different performance
	on different levels of transfer operations.
	To answer this question from another angle,
	we further look into the actual topics trained from \softlink{} and \voclink{}
	in this section.
	The general idea is to look into the same topic output from \softlink{} and \voclink{}
	and see what topic words they have in common (denoted as $\mathcal{W}^+$),
	and what words they have exclusively,
	denoted as $\mathcal{W}^{-,\textsc{soft}}$ and $\mathcal{W}^{-,\textsc{voc}}$
	for \softlink{} and \voclink{} respectively.
	The words in $\mathcal{W}^{-,\textsc{voc}}$ are those with lower
	topic coherence
	and are thus the key to understanding the suboptimal performance of \voclink{}.

	\subsubsection{Aligning Topics}
	To this end, the first step is to align possible topics between \voclink{} and \softlink{},
	since the initialization of Gibbs samplers is random.
	Let $\{\mathcal{W}^{\textsc{voc}}_k\}_{k=1}^K$ and $\{\mathcal{W}^{\textsc{soft}}_k\}_{k=1}^K$
	be the $K$ topics learned by \voclink{} and \softlink{} respectively,
	from the same training conditions.
	For each topic pair $(k,k')$
	we calculate the Jaccard index $\mathcal{W}_k^{\textsc{voc}}$ and $\mathcal{W}_{k'}^{\textsc{soft}}$,
	one for each language,
	and use the average over the two languages as the matching score $m_{k,k'}$ of the topic pair:
	\begin{align}
		m_{k,k'}~=~\frac{1}{2}\bigg( J\left(\mathcal{W}_{k,\ell_1}^{\textsc{voc}},\mathcal{W}_{k',\ell_1}^{\textsc{soft}}\right) + J\left(\mathcal{W}_{k,\ell_2}^{\textsc{voc}},\mathcal{W}_{k',\ell_2}^{\textsc{soft}}\right)\bigg),
	\end{align}
	where $J(X,Y)$ is the Jaccard index between sets $X$ and $Y$.
	Thus, there are $K^2$ matching scores with a number of topics $K$.
	We set a threshold of $0.8$,
	so that a matching score is valid only when it is greater than $0.8\cdot \max m_{k,k'}$ over all the $K^2$ scores.
	For each topic $k$,
	if its matching score is valid,
	we align $\mathcal{W}_k^{\textsc{voc}}$ with $\mathcal{W}_{k'}^{\textsc{soft}}$,
	and treat them as potentially the same topic.
	When multiple matching scores are valid,
	we use the topic with highest score and ignore the rest.

	\subsubsection{Comparing Document Frequency}
	
	Using the approximate alignment algorithm we described above,
	we are now able to compare each aligned topic pair between \voclink{} and \softlink{}.
	
\begin{figure}
		\centering
		\includegraphics[width=\linewidth]{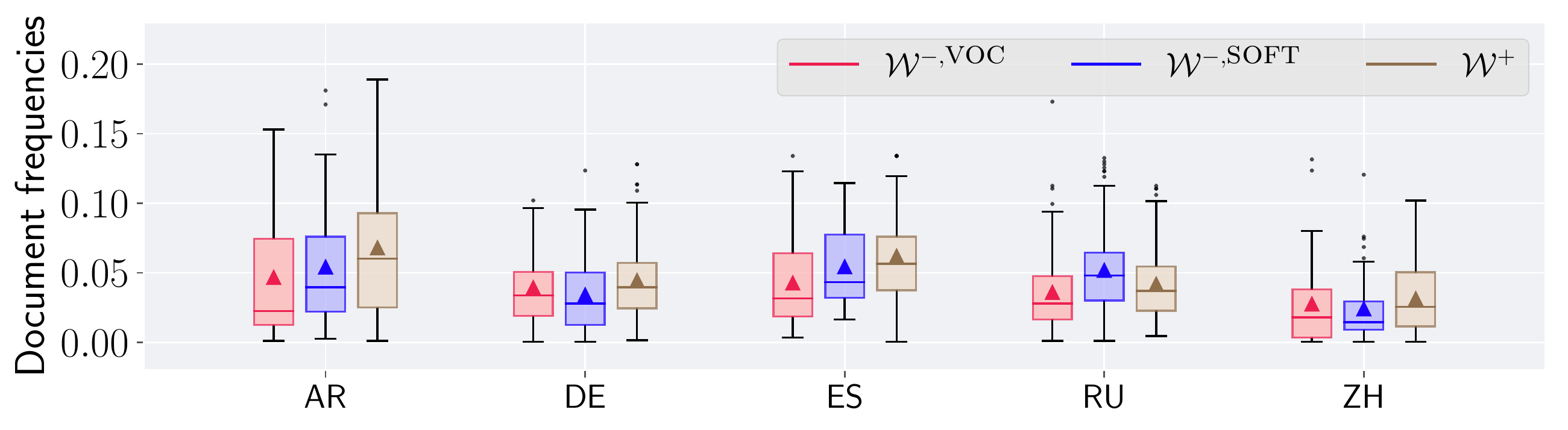}
		\includegraphics[width=\linewidth]{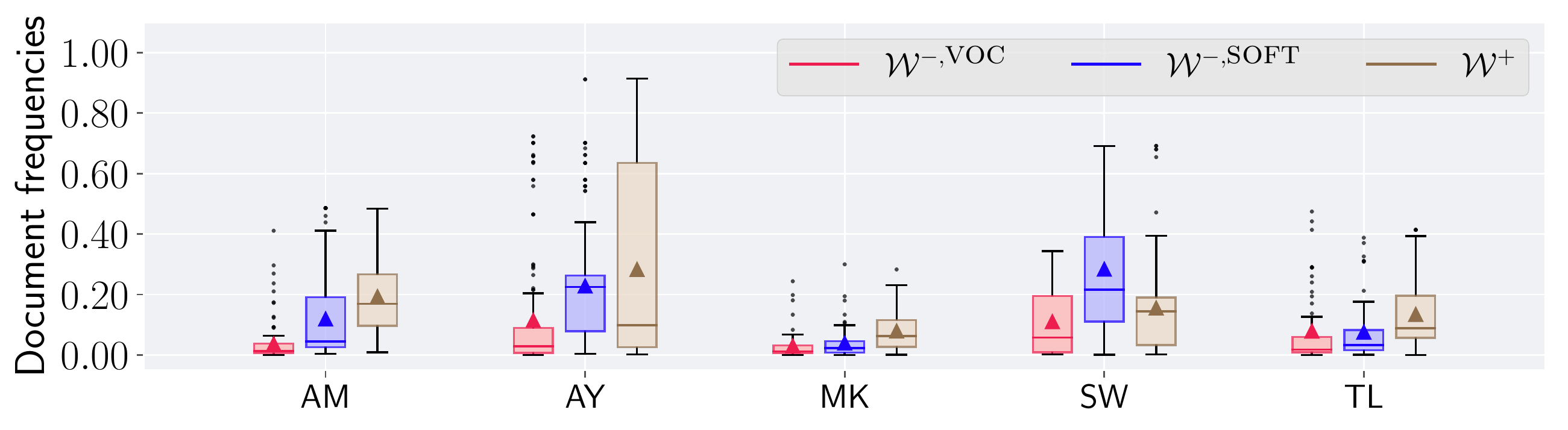}
		\caption{
			Average document frequencies of $\mathcal{W}^{-,\textsc{voc}}$
			are generally lower than $\mathcal{W}^{-,\textsc{soft}}$
			and $\mathcal{W}^{+}$, shown in the triangle markers.
		}
		\label{fig:entropybarchart}
	\end{figure}

	For a word type $w$, we define the document frequency as the percentage of documents where $w$ appears.
	A low document frequency of word $w$ implies that $w$ only appears in a small number of documents.
	For every aligned topic pair $\left( \mathcal{W}_i, \mathcal{W}_j \right)$
	where $\mathcal{W}_i$ and $\mathcal{W}_j$ are topic word sets from \softlink{} and \voclink{} respectively,
	we have three sets of topic words derived from this pair: 
	\begin{align}
		\mathcal{W}^{+\phantom{,soft}} ~=~& \mathcal{W}_i \cap \mathcal{W}_j,\\
		\mathcal{W}^{-,\textsc{voc\phantom{s}}} ~=~& \mathcal{W}_i~\backslash~\mathcal{W}_j,\\
		\mathcal{W}^{-,\textsc{soft}} ~=~& \mathcal{W}_j~\backslash~\mathcal{W}_i.
	\end{align}
	Then we calculate the average document frequencies over all the words in each of the sets,
	and we show the results in \fig{fig:entropybarchart}.
	
	We observe that the average document frequencies over words in $\mathcal{W}^{-,\textsc{voc}}$
	are consistently lower in every language,
	while those in $\mathcal{W}^{+}$ are higher.
	This implies that \voclink{} tends to give rare words higher probability in the topic-word distributions.
	In other words, \voclink{} gives high probabilities to words
	that only appear in specific contexts, such as named entities.
	Thus, when evaluating topics using a reference corpus,
	the co-occurrence of such words with other words is relatively low
	due to lack of that specific context in the reference corpus.
	
\begin{figure}
		\centering
		\includegraphics[width=\linewidth]{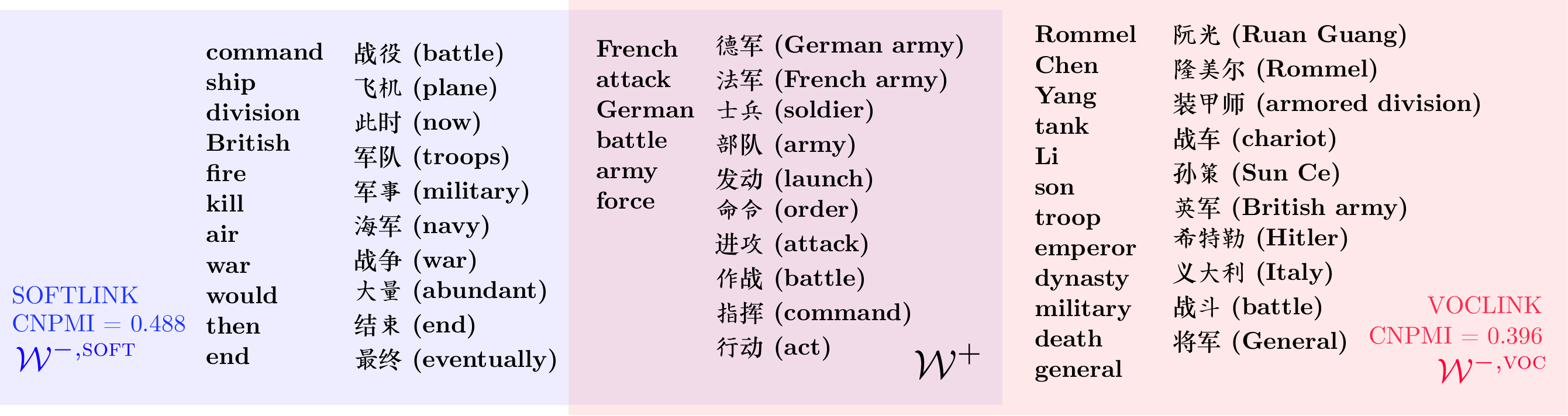}
		\caption{An example of real data showing the topic words of \softlink{} and \voclink{}.
			Words that appear in both models are in $\mathcal{W}^+$;
			words that only appear in \softlink{} or \voclink{} are included in $\mathcal{W}^{-,\textsc{soft}}$ or $\mathcal{W}^{-,\textsc{voc}}$, respectively.}
		\label{fig:topicexample}
	\end{figure}

	We show an example of an aligned topic in \fig{fig:topicexample}.
	In this example,
	we see that although both \voclink{} and \softlink{}
	can discover semantically coherent words shown in $\mathcal{W}^{+}$,
	\voclink{} focuses more on words that only appear in
	specific contexts:
	there are many words (mostly named entities) in $\mathcal{W}^{-,\textsc{voc}}$
	that only appear in one document.
	Due to lack of this very specific context in the reference corpora,
	the co-occurrence of these words with other more general words
	is likely to be zero, resulting in lower \cnpmi{}.

	\subsection{Comparing Transfer Strength}\label{sec:comparestrength}
	
	While we have looked at the topics to explain what kind of words produced by \voclink{}
	make the model's performance lower than \softlink{},
	in this section,
	we try to explain why this happens
	by analyzing their transfer operations.
	Recall that \voclink{} defines transfer operation on topic-node distributions $\{\phi_{k,r}\}_{k=1}^K$ (Equation~(\ref{voctr})),
	while \softlink{} defines transfer on document-topic distributions $\theta$.
	The difference of transfer levels with same resource
	leads to a suspicion that document level has a ``stronger'' transfer power.

	The first question is to understand how this transfer operation
	actually functions in the training of topic models.
	During Gibbs sampling of monolingual LDA,
	the conditional distribution for a token, denoted as $\mathcal{P}$
	is calculated by conditioning on all the other tokens and their topics,
	and can be factorized into two conditionals---document level $\mathcal{P}_\theta$
	and word level $\mathcal{P}_\phi$.
	Let the current token be of word type $w$,
	and $\mathbf{w}_-$ and $\mathbf{z}_-$ all the other words and their current topic assignments in the corpus.
	The conditional is then
	\begin{align}
		\mathcal{P}_k~=~&\Pr\left( z = k | w, \mathbf{w}_-, \mathbf{z}_- \right)\\
		~\propto~& \left( n_{k|d} + \alpha_k \right)
		\cdot \frac{n_{w|k} + \beta_w}{n_{\cdot|k} + \mathbf{1}^\top\beta }\\
		~=~&\mathcal{P}_{\theta k} \cdot \mathcal{P}_{\phi k},
	\end{align}
	where $n_{k|d}$ is the number of topic $k$ in document $d$,
	$n_{w|k}$ the number of word type $w$ in topic $k$,
	$n_{\cdot | k}$ the number of tokens assigned to topic $k$,
	and $\mathbf{1}$ an all-one vector.
	In this equation,
	the final conditional distribution can be treated as a ``vote'' from the two conditionals, \textit{i.e.,} $\mathcal{P}_{\theta}$ and $\mathcal{P}_\phi$~\cite{YuanGHDWZXLM15}.
	If $\mathcal{P}_\phi$ is a uniform distribution,
	then $\mathcal{P} = \mathcal{P}_\theta$,
	meaning the conditional on document $\mathcal{P}_\theta$
	dominates the decision of choosing a topic,
	while the conditional on word $\mathcal{P}_\phi$ is uninformative.
	
	We apply this similar idea to multilingual topic models.
	For a token in language $\ell_2$, we let $w$ be its word type,
	and $\mathcal{P}$ can also generally be factorized to two individual conditionals,
	\begin{align}
		\mathcal{P}_k~=~&\Pr\left( z = k | w, \mathbf{w}_-, \mathbf{z}_- \right)\\
		~\propto~& \underbrace{\left[ n_{k|d} + h_\theta\left( \delta,\mathbf{N}^{(\ell_1)},\alpha \right)_k \right]}_{\mathcal{P}_{\textsc{doc},k}}
		\cdot \underbrace{\frac{n_{w|k} + h_\phi\left( \delta',\mathbf{N}^{(\ell_1)},\beta \right)_w}{n_{\cdot|k} + \mathbf{1}^\top h_\phi\left( \delta',\mathbf{N}^{(\ell_1)},\beta \right)}}_{\mathcal{P}_{\textsc{voc},k}} \label{eq:tq}\\
		~=~&\mathcal{P}_{\textsc{doc},k} \cdot \mathcal{P}_{\textsc{voc},k},
	\end{align}
	where the transfer operation is clearly incorporated into the calculation of the conditional,
	and $\mathcal{P}_{\textsc{doc}}$ and $\mathcal{P}_{\textsc{voc}}$
	are conditional distributions on document and word levels respectively.
	Thus,
	it is easy to see how transfers on different levels contribute to the decision of a topic.
	This is also where our comparison of ``transfer strength'' starts.

\begin{figure}
		\centering
		\includegraphics[width=\linewidth]{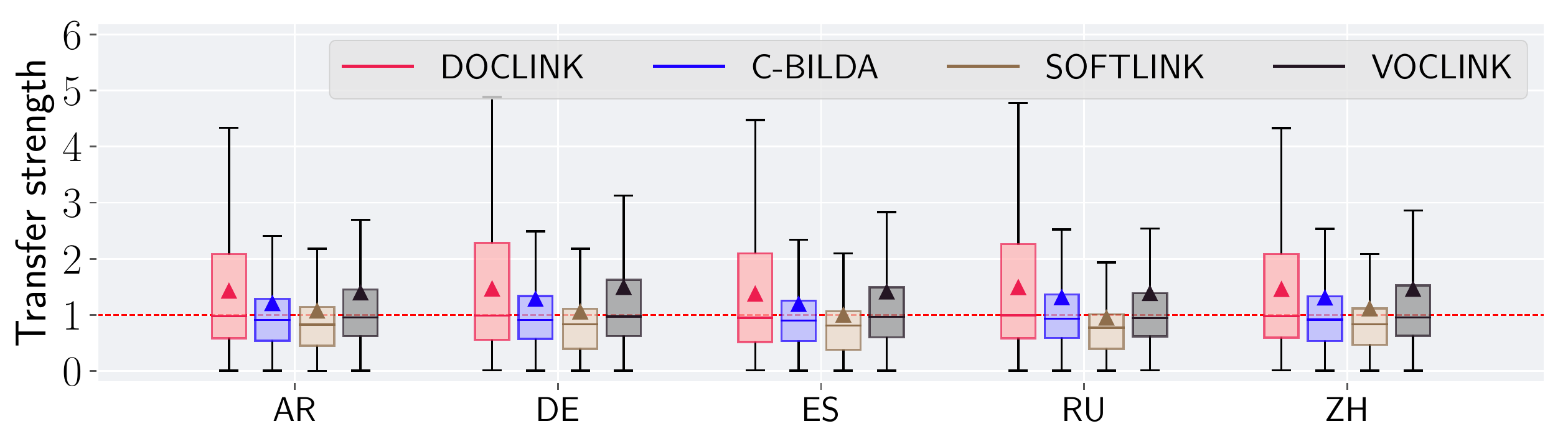}
		\includegraphics[width=\linewidth]{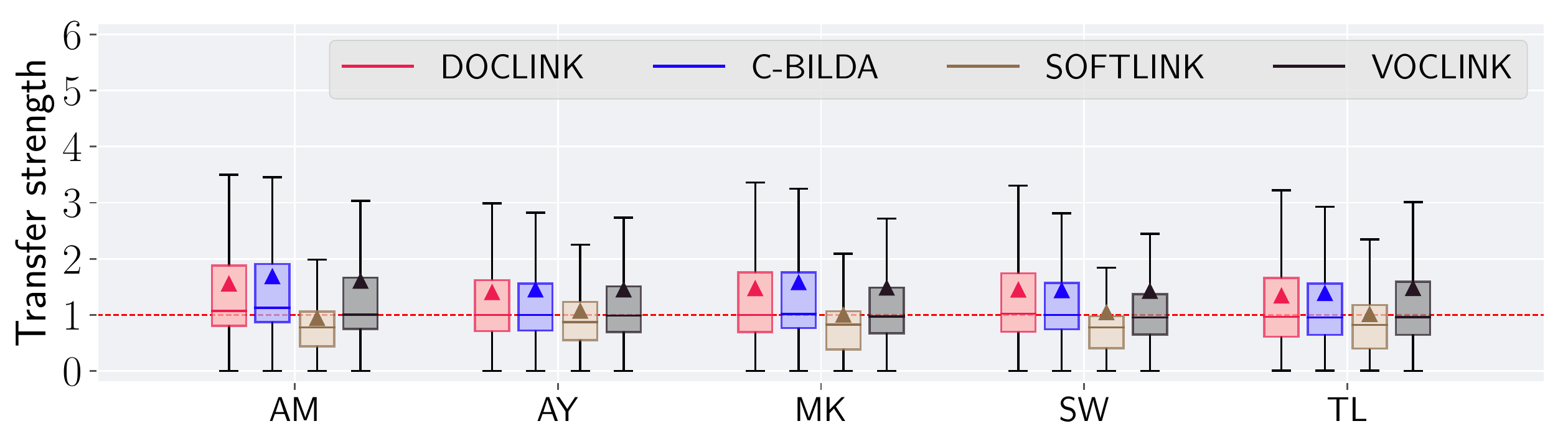}
		\caption{Comparisons of transfer strength. A value of one (shown in red dot line) means an equal balance of transfer between document and word levels. We notice \softlink{} has the most balanced transfer strength, while \voclink{} has stronger transfer at the document level although its transfer operation is defined on the word level.}
		\label{fig:strength}
	\end{figure}

	To apply this idea, for each token, we first obtain three distributions described before:
	$\mathcal{P}$, $\mathcal{P}_{\textsc{doc}}$, and $\mathcal{P}_{\textsc{voc}}$.
	Then we calculate cosine similarities $\cos\left(\mathcal{P}_{\textsc{doc}}, \mathcal{P}\right)$
	and $\cos\left(\mathcal{P}_{\textsc{voc}}, \mathcal{P}\right)$.
	If $r=\frac{\cos\left(\mathcal{P}_{\textsc{doc}}, \mathcal{P}\right)}{\cos\left(\mathcal{P}_{\textsc{voc}}, \mathcal{P}\right)} > 1$,
	we know that $\mathcal{P}_{\textsc{doc}}$ is dominant and helps shape the conditional distribution $\mathcal{P}$;
	in other words, the document level transfer is stronger.
	We calculate the ratio of similarities $r=\frac{\cos\left(\mathcal{P}_{\textsc{doc}}, \mathcal{P}\right)}{\cos\left(\mathcal{P}_{\textsc{voc}}, \mathcal{P}\right)}$
	for all the tokens in every model,
	and take the model-wise average over all the tokens (\fig{fig:strength}).
	The most balanced situation is $r=1$,
	meaning transfers on both word and document levels
	are contributing equally to the conditional distributions.

	From the results, we notice that both \doclink{} and \cbilda{}
	have stronger transfer strength on the document level,
	which means that the transfer operations on the document levels are
	actually informing the decision of a token's topic.
	However, we also notice that \voclink{} has very comparable transfer strength
	to \doclink{} and \cbilda{},
	which makes less sense,
	since \voclink{} defines transfer operations on the word level.
	This implies that transferring knowledge on the word level is weaker.
	This also explains why, in the previous section,
	\voclink{} tends to find topic words appearing in only a few documents.
	
	It is also interesting to see \softlink{} having a relatively good balance
	between document and word levels,
	with consistently the most balanced transfer strengths across all models and languages.

	\section{Remarks and Conclusions}
	
	Multilingual topic models use corpora in multiple languages as input with additional language resources
	as supervision.
	The traits of these models inevitably lead to a wide variety of training scenarios,
	especially when a language's resources are scarce,
	while most previous
	studies on multilingual topic models
	have not analyzed in depth the appropriateness of different models for different training situations and resource availability.
	For example, experiments are most often done in European languages,
	with models that are typically trained on parallel or comparable corpora.
	
	The contributions of our study are to provide a unifying framework of these different models,
	and to systematically analyze their efficacy in different training situations.
	We conclude by summarizing our findings along two dimensions---training corpora characteristics and dictionary characteristics,
	since these are the necessary components to enable crosslingual knowledge transfer.
	
	\subsection{Model Selection}
	Document-level models are shown to work best when the corpus
	is parallel or at least comparable.
	In terms of learning high-quality topics,
	\doclink{} and \cbilda{} yield very similar results.
	However, since \cbilda{} has a ``language selector'' mechanism in the generative process,
	it is slightly more efficient for training Wikipedia articles in low-resource languages,
	where the document lengths have large gaps compared to English.
	\softlink{}, on the other hand,
	only needs a small dictionary to enable document-level transfer,
	and yields very competitive results.
	This is especially useful for low-resource languages
	when the dictionary size is small
	and only a small number of comparable document pairs are available for training.
	
	Word-level models are harder to achieve full potential of transfer,
	due to limits in the dictionary size and training sets, and unrealistic assumptions of the generative process regarding dictionary coverage.
	The representative model, \voclink{},
	has similarly good performance on document classification as other models,
	but the topic qualities according to coherence-based metrics are lower.
	Comparing to \softlink{}, which also requires a dictionary as resource,
	directly modeling word translations in \voclink{}
	turns out to be a less efficient way of transferring dictionary knowledge.
	Therefore, 
	when using dictionary information,
	we recommend \softlink{} over \voclink{}.
	
	\subsection{Relations to Other Crosslingual Representations}
	
	As an alternative method to learning crosslingual representations,
	crosslingual and domain agnostic embeddings have been gaining attention~\cite{HuangLCWSB18,Ruder17,UpadhyayFDR16}.
	Similar to the topic space in multilingual topic models,
	crosslingual embeddings learn semantically consistent features
	in a shared embedding space for all languages.
	A very common strategy for learning crosslingual embeddings
	is to use a projection matrix as supervision~\cite{VulicK16,FaruquiD14}
	or sub-objective~\cite{TsvetkovD16,DinuB14}
	to learn a common embedding space for separately trained monolingual embeddings.
	
	In multilingual topic models,
	the supervision matrix $\delta$ plays the role of a projection matrix between languages.
	For example,
	in \doclink{},
	$\delta_{d_{\ell 2},d_{\ell 1}}$ projects document $d_{\ell_2}$
	to the document space of $\ell_1$ (\eq{ident}).
	\softlink{} provides a simple extension
	by forming $\delta$ to a matrix of transfer distirbutions
	based on word-level document similarities.
	\voclink{} applies projections in the form of
	word translations.
	
	Thus, we can see that the formation of projection matrices in multilingual topic models
	is still static, \textit{i.e.,} used as supervision,
	and restricted to an identity matrix or a simple pre-calculated matrix.
	A generalization would be to add learning the projection matrix itself
	as an objective into multilingual topic models.
	This could be a way for improving \voclink{}
	by extending word associations to polysemy across languages,
	and making it less dependent on context.

	\subsection{Future Directions}
	
	Our study inspires future work in two directions.
	The first direction is to increase the efficiency of word-level knowledge transfer.
	For example, it is possible to use co-location information of translated words
	to transfer knowledge, though cautiously, to untranslated words.
	It has been shown that word-level models can help find new word translations, for example
	by using the existing dictionary as ``seed'',
	and gradually adding more internal nodes to the tree structure using trained topic-word distributions.
	Additionally,
	our analysis showed the benefits of using a ``language selector'' in \cbilda{}
	to make the generative process of \doclink{} more realistic,
	and one could also implement a similar mechanism in \voclink{} to
	make the conditional distributions for tokens less dependent on specific context.
	
	The second direction is more general.
	By systematically synthesizing various models
	and abstracting the knowledge transfer mechanism through an explicit transfer operation,
	we can construct models that shape the probabilistic distributions of a target language
	using that of a source language.
	By defining different transfer operations,
	more complex and robust models can be developed,
	and this transfer formulation may provide new ways of constructing models than with a traditional joint formulation.
	For example, \softlink{} is generalization \doclink{}
	based on transfer operations
	that does not have an equivalent joint formulation.
	This framework for thinking about multilingual topic models
	may lead to new ideas for other models.

	\newpage
	\starttwocolumn

\end{document}